%% file: main.tex
\pdfoutput=1
\documentclass[11pt]{article}

\usepackage[]{acl}

\usepackage{times}
\usepackage{latexsym}
\usepackage{graphicx}
\usepackage{adjustbox}
\usepackage{multirow}
\usepackage[normalem]{ulem}
\usepackage{xcolor,colortbl}
\usepackage{soul}
\setuldepth{Berlin}
\usepackage{paralist}
\usepackage{xfrac}
\usepackage[inline]{enumitem}
\usepackage{graphicx}
\usepackage{url}
\usepackage{textcomp}
\usepackage{caption}
\usepackage{subcaption}
\usepackage{float}
\usepackage{placeins}
\usepackage{aliascnt}
\usepackage{mathtools}
\usepackage{amsmath}
\usepackage{hhline}
\usepackage{booktabs}
\usepackage{soul}
\usepackage{bbm}
\usepackage{amssymb}
\usepackage{pifont}
\usepackage{adjustbox}
\usepackage{array}
\usepackage{lipsum}
\usepackage{scrextend}
\newcommand\setalign{4pt}
\usepackage[percent]{overpic}
\usepackage{titlesec}
\usepackage{setspace}
\usepackage{hyperref}
\usepackage{microtype}  % Ensure microtype is loaded last
\usepackage{footnote}

\usepackage[T1]{fontenc}
\usepackage[utf8]{inputenc}

\newcommand\blfootnote[1]{%
  \begingroup
  \renewcommand\thefootnote{}\footnote{#1}%
  \addtocounter{footnote}{-1}%
  \endgroup
}

\title{Space Decomposition for Sentence Embedding}

\author{
Wuttikorn Ponwitayarat\textsuperscript{*\dag}, Peerat Limkonchotiwat\textsuperscript{*\dag},\\
\textbf{Ekapol Chuangsuwanich}\textsuperscript{‡}, \textbf{Sarana Nutanong}\textsuperscript{\dag}\\
  \textsuperscript{\dag}School of Information Science and Technology, VISTEC, Thailand\\
  \textsuperscript{‡}Department of Computer Engineering, 
  Faculty of Engineering, \\ Chulalongkorn University, Thailand \\
  \texttt{\{wuttikorn.p\_s22,peerat.l\_s19,snutanon\}@vistec.ac.th}
  \\
    \texttt{ekapolc@cp.eng.chula.ac.th}
  }

\begin{document}
\maketitle
\blfootnote{\textsuperscript{*}Equal contributions}
\begin{abstract}
\input{0_abstract}
\end{abstract}

\input{1_introduction}
\input{2_related_work}

\input{3_Methodology}

\input{4_Exp_setup}
\input{5_Exp_results}

\input{6_Conclusion}

\bibliography{anthology,custom}
\bibliographystyle{acl_natbib}

\appendix
\input{7_Appendix}

\end{document}

%% file: 0_abstract.tex
Determining sentence pair similarity is crucial for various NLP tasks.
A common technique to address this is typically evaluated on a continuous \emph{semantic textual similarity} scale from 0 to 5. 
However, based on a linguistic observation in STS annotation guidelines, we found that the score in the range [4,5] indicates an upper-range sample, while the rest are lower-range samples.
This necessitates a new approach to treating the upper-range and lower-range classes separately.  
In this paper, we introduce a novel embedding space decomposition method called \emph{MixSP} utilizing a \emph{Mixture of Specialized Projectors}, designed to distinguish and rank upper-range and lower-range samples accurately. 
The experimental results demonstrate that MixSP decreased the overlap representation between upper-range and lower-range classes significantly while outperforming competitors on STS and zero-shot benchmarks.~\footnote{The code and models are available at \url{https://github.com/KornWtp/MixSP}.}

%% file: 1_Introduction.tex
\section{Introduction}
\label{sec:intro}

Determining the similarity between sentence pairs is fundamental to many downstream applications such as text classification, search, and ranking. 
Usually, sentence pair similarity is assessed via \emph{Semantic Textual Similarity (STS)}, where each sample contains a pair of sentences, and their label denotes the degree of similarity, which uses scores from 0 to 5, where 5 represents the highest degree of similarity.
Studies have shown that improving the ability to rank sentence pairs according to their similarities enhances text classification accuracy~\cite{DBLP:conf/emnlp/GaoYC21,limkonchotiwat-etal-2022-congen,miao2024enhancing} and reranking mean-average precision~\cite{wang-etal-2021-tsdae-using}.

\begin{figure}[t]
\centering
% \begin{subfigure}[b]{0.5\textwidth}
% \vspace{-2mm}
% \centering
% \includegraphics[width=\textwidth]{images/ridgeplot.Gold-BERT-Base.sts-cd_test_test-set.pdf}
% \label{fig:gold_overlap_area_cd_test}
% \end{subfigure}
% \hfill
\begin{subfigure}[b]{0.48\textwidth}
\vspace{-4mm}
\centering
\includegraphics[width=1\textwidth]{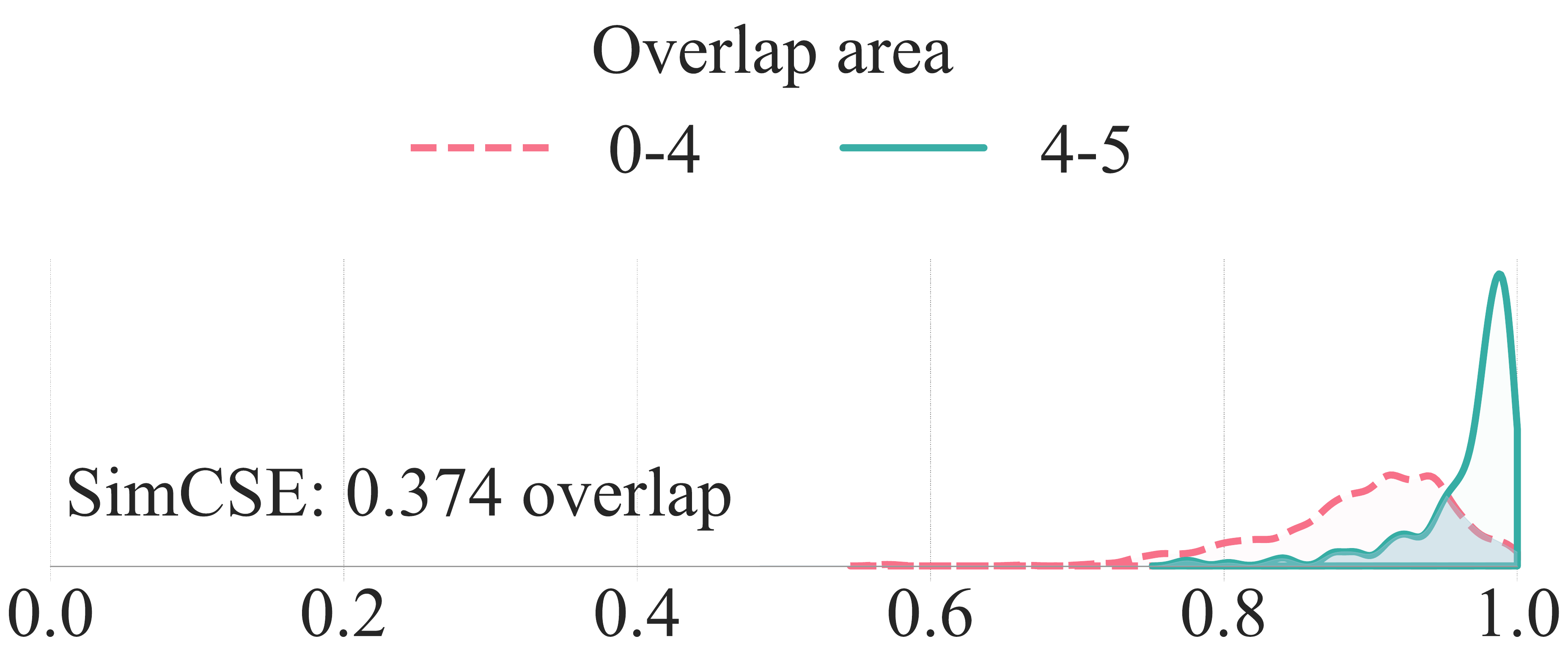}
\label{fig:simcse_overlap_area_cd_test}
\end{subfigure}
\hfill
\begin{subfigure}[b]{0.48\textwidth}
\vspace{-4mm}
\centering
\includegraphics[width=1\textwidth]{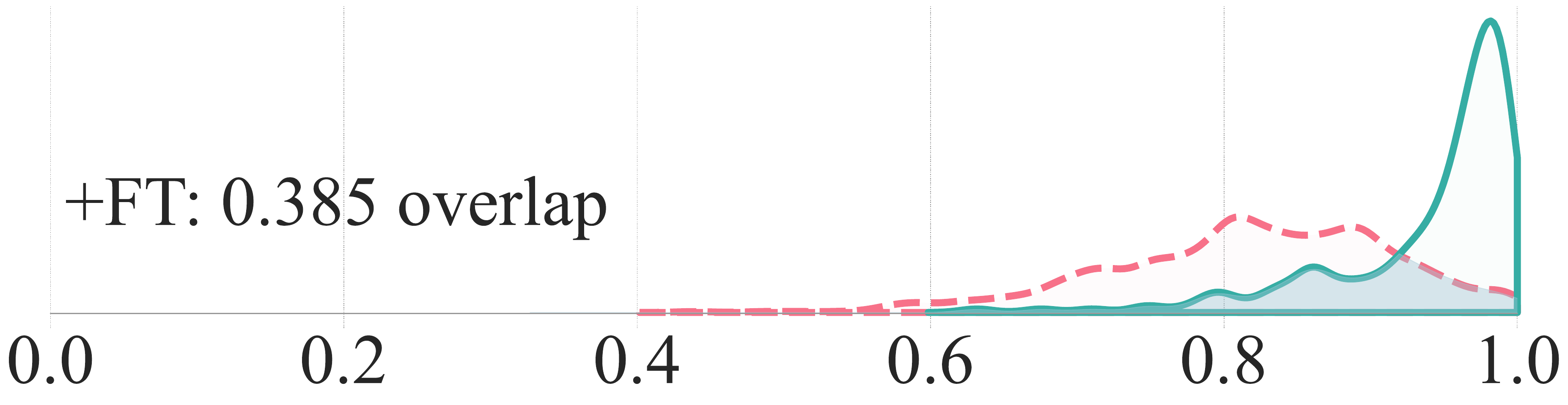}
\label{fig:simcse_ft_overlap_area_cd_test}
\end{subfigure}
\begin{subfigure}[b]{0.48\textwidth}
\vspace{-4mm}
\centering
\includegraphics[width=1\textwidth]{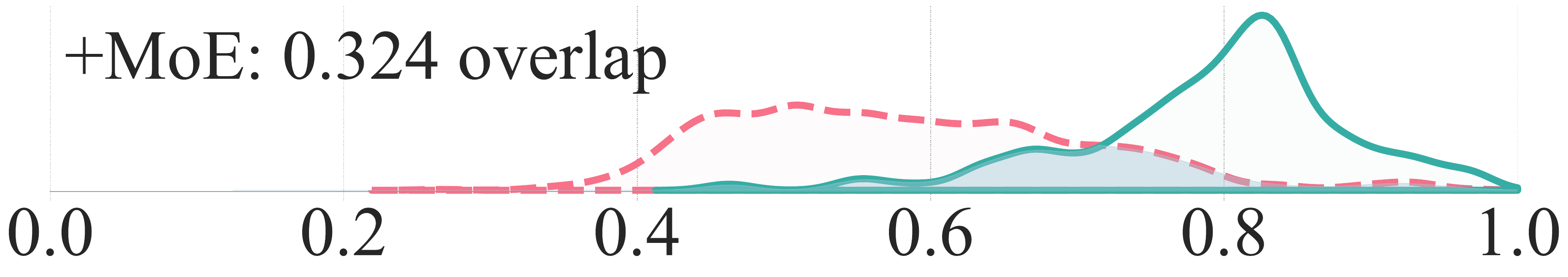}
\label{fig:simcse_moe_overlap_area_cd_test}
\end{subfigure}
\begin{subfigure}[b]{0.48\textwidth}
\vspace{-4mm}
\centering
\includegraphics[width=1\textwidth]{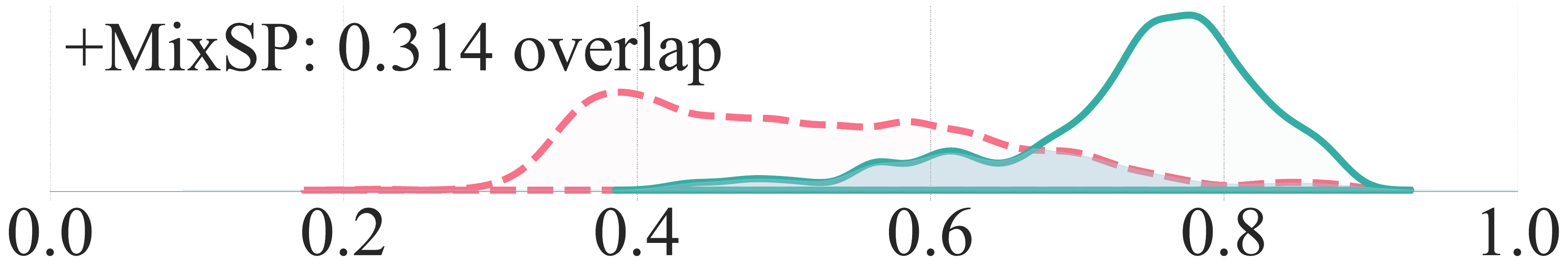}
\label{fig:simcse_mixsp_overlap_area_cd_test}
\end{subfigure}
\vspace{-8mm}
\caption{Cosine similarity distributions for BERT-Base formulated on Gaussian estimation. The overlap value refers to the intersection between the upper and lower ranges. We use the data from the CDSC-R test data.}
\vspace{-6mm}
\label{fig:overlap_area_cd_test}
\end{figure}

A common approach to solving the STS problem is employing a pre-trained language model~\cite{DBLP:conf/naacl/DevlinCLT19,DBLP:journals/corr/abs-1907-11692} and finetuning it with a supervised learning objective.
%
% This technique involves mapping sentences into an embedding space, where they can be effectively separated based on their degree of similarity, as evaluated through the Semantic Textual Similarity (STS) task. 
%
In particular, we aim to construct an embedding space in which the cosine similarity between sentence pairs reflects the degree of similarity; contrastive learning is a popular method to achieve such an embedding space~\cite{yan-etal-2021-consert, DBLP:conf/emnlp/GaoYC21, DBLP:conf/emnlp/JiangJHZWZWHDZ22, wang2022diffaug}.
Regardless of the differences in their data augmentation strategies, all aforementioned methods treat the degree of similarity as a continuous range.
In other words, all learning methods indifferently treat sentence pairs with different degrees of similarities within the STS score range [0,5].

In this paper, we challenge the common practice of treating STS scores as a continuous spectrum. 
Several studies observed that the score range [4,5] signifies semantically related (i.e., upper-range) samples, while the rest represents unrelated (i.e., lower-range) samples~\cite{DBLP:conf/emnlp/GaoYC21, chuang2022diffcse}.
Consequently, the STS problem should be considered a ranking within two distinct classes rather than one continuous spectrum.

We introduce a novel embedding space decomposition method called \emph{MixSP} utilizing a \emph{Mixture of Specialized Projectors}. 
%
%Our proposed method, \emph{MixSP}, is a \emph{Mixture of Specialize Projectors} for Sentence Embedding. 
%
The novelty of MixSP lies in a carefully designed learning pipeline with the following traits: (i) the ability to distinguish upper-range from lower-range samples and (ii) the ability to accurately rank sentence pairs within each class. 
In particular, our method uses a routing network and two specialized projectors to handle upper-range and lower-range representations, resulting in a better STS performance overall.

Figure~\ref{fig:overlap_area_cd_test} illustrates how our embeddings can better distinguish different sentence pairs as compared to our competitors: SimCSE~\cite{DBLP:conf/emnlp/GaoYC21}, FT~\cite{reimers-2019-sentence-bert}, and MoE~\cite{NEURIPS2022_2f00ecd7} compared to that of MixSP. 
We quantify the confusion between upper-range and lower-range classes as the cosine score overlaps between these two classes using Gaussian kernel density estimates. 
A smaller overlap indicates the ability to distinguish the upper-range and lower-range classes.
We can see that MixSP obtains the smallest overlap between upper-range and lower-range classes of 31.4\% while all competitors have an overlap ranging from 32.4\% to 38.5\%.
Regarding the similarity ranking performance, our method also produces superior performance compared to these competitors (as shown in Figure~\ref{fig:overlap}, Section~\ref{subsec:why}).
These results demonstrate that our method improves the ability to distinguish upper-range and lower-range samples and rank sentence pairs according to their similarities. 

Our contributions are as follows:
\begin{compactitem}[\hspace{\setalign}•]
    \item We have recast the sentence embedding paradigm from one embedding space containing upper-range and lower-range to separate embedding space for each group.

    \item We propose a novel embedding space decomposition technique called \emph{Mixture of Specialized Projectors} (\emph{MixSP}). 
    Our model has the ability to distinguish upper-range and lower-range samples while accurately ranking sentence pairs within each class.
        
    %\item We propose a new embedding evaluation by evaluating the overlap area between positive and negative samples based on the STS score and Gaussian estimation. 
    %
    %This new evaluation highlights the performance gap in sentence embedding works.

    % \item We experiment with six studies, four methods, and 14 datasets to confirm the claim of our contributions\footnote{We release code and model up to the acceptance.}.
    
    % \item Our experiments revealed significant reductions in computational costs, training time, and memory usage, demonstrating the practical efficiency of our approach across various tasks and datasets.\footnote{We release code and model up to the acceptance.}

    \item We demonstrate the efficiency of our method on STS and zero-shot benchmarks. In addition, we provide deep analyses of (i) performance efficiency and (ii) design choice in embedding space decomposition settings. 
    
    % Our experiments and analysis in this study demonstrate the efficiency and promising direction of challenging the sentence embedding problem from one continuous embedding space to multiple embedding spaces. 

    % Moreover revealed significant reductions in computational costs, training time, and memory usage, demonstrating the practical efficiency of our approach across various tasks and datasets.\footnote{We release code and model up to the acceptance.}
    
\end{compactitem}

%% file: 2_related_work.tex
\section{Related Work}

\subsection{Sentence Representation}
Currently, researchers typically use pre-trained language models and supervised contrastive learning to train sentence representation models. 
The main goal of contrastive learning is to maximize the similarity between anchor and positive while minimizing the similarity between anchor and negative.
The key component of contrastive learning is data augmentation for positive and negative pairs.
\citet{DBLP:conf/emnlp/GaoYC21} proposed SimCSE, a contrastive learning for sentence embedding. 
SimCSE used dropout masks in two forward passes as the data augmentation method.

\citet{DBLP:conf/emnlp/JiangJHZWZWHDZ22} proposed PromptBERT, a prompt-based sentence embedding.
PromptBERT used contrastive learning with template denoising to generate positive pairs, while negative pairs are sentences within the same mini-batch.

\citet{wang2022diffaug} proposed DiffAug, a two-stage training objective. 
%
% The training objective of DiffAug is similar to SimCSE, but DiffAug used a pre-fix model to produce a hard positive. 
%
The training objective of DiffAug is similar to SimCSE, but DiffAug used a contextual prompt to produce a hard positive, improving the generalizability of the embedding~space.

Additionally, other works used various augmentation to obtain augmented texts.
Notable techniques include back-translation~\cite{fang2020cert,limkonchotiwat-etal-2022-congen, limkonchotiwat-etal-2023-sct}, MLM~\cite{yang-etal-2021-universal, chuang2022diffcse}, and prompting~\cite{zhou-etal-2022-flipda, wang-etal-2022-promda, DBLP:conf/emnlp/JiangJHZWZWHDZ22, jiang-etal-2022-improved}.

%%%% SOTA ดีนะ แต่มันมี room of improvement อยู่ คือมัน fail to dintingush positive and negative samples
%%%% เราอยากแยก positive and negative เพื่อจะทำให้ผลดีขึ้น

% These works demonstrate a high score on STS benchmarks. 
% %
% However, based on our overlap evaluation, we found that current works fail to distinguish positive and negative samples. 
% % 
% We require a new method to address this problem. 

\begin{figure*}[t]
% \hspace*{-4mm}
\centering
\includegraphics[width=1\textwidth]{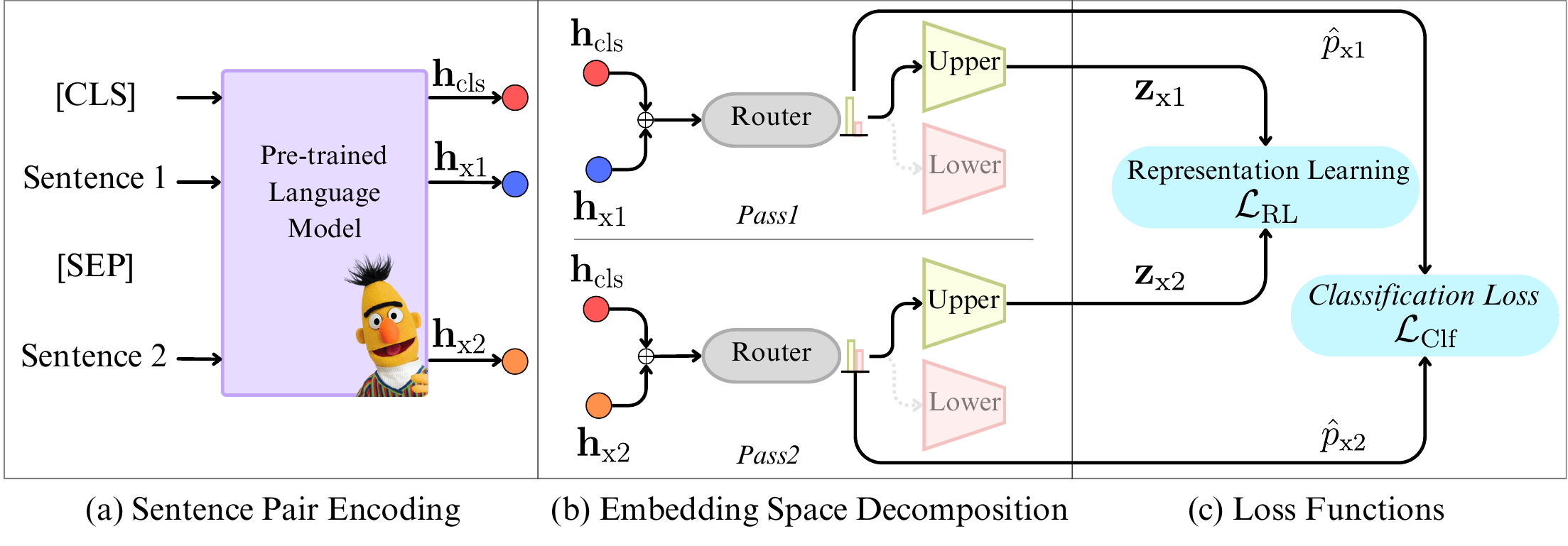}
\vspace{-3mm}
\caption{The overview of \emph{Mixture of Specialized Projectors (MixSP)}. (a) Given an upper-range sample, we encode the sample with a pre-trained language model. (b) We use a router to classify a class of sentences 1 and 2 (upper-range or lower-range). The final representation is formulated by projecting the representation with specialized projectors. (c) We improve the classification and representation with our training losses $\mathcal{L}_{\text{Clf}}$ and $\mathcal{L}_{\text{RL}}$, respectively. }
\vspace{-4mm}
\label{fig:overview}
\end{figure*}

\subsection{Embedding Space Decomposition} %% EoE: ensemble of expertise
%
% \citet{DBLP:conf/icnn/SzymanskiL93, DBLP:journals/neco/JordanJ94} proposed a Mixture-of-Experts (MoE), the goal of MoE is to distinguish samples in the same embedding with different experts.
% %
% MoE comprises two main components: \emph{routing} and \emph{expertise} networks.
% %
% The routing network performs a classification of experts' weight, as a softmax score. 
% %
% Each expertise produces representation vectors from the input weighted with the softmax score.
% %
% The final representation is acquired by aggregating the vectors from the experts.

Embedding space decomposition is the task of partitioning data into distinct subsets within the embedding space, enhancing model performance and understanding through focused representations.
A common technique is to separate the space with semantic features.
\citet{DBLP:conf/icpr/WangCWK20} proposed semantic subspace analysis to break down the high-dimensional embedding space into semantic groups and examine their interrelationships. 
%
% Considering intra-group and inter-group descriptors, it crafts an efficient sentence representation capturing the semantic details of its constituent words.
%
\citet{DBLP:conf/ijcnlp/OpitzF22} decomposed the embedding space to unveil interpretable semantic features within sentence embeddings, targeting semantic roles, negation, and quantification for a deeper understanding of the conveyed meaning.

Recently, researchers employed Mixture-of-Experts (MoE), which partitions the space into smaller subspaces managed by specialized experts, to handle varied data aspects~\cite{DBLP:conf/iclr/LiSYWRCZ023,DBLP:conf/cvpr/ChenSDCZLG23,DBLP:conf/icml/ChowdhuryZW0C23}. 
The key components of MoE comprise \emph{routing networks} that classify embedding space weights and aggregate embedding spaces from \emph{expertise networks} to formulate the final representation.
%
% In particular, MoE used in image embedding, as demonstrated by GMoE   for attribute recognition, Mod-Squad \citet{DBLP:conf/cvpr/ChenSDCZLG23} for multitasking and expert-task optimization, and \citet{DBLP:conf/icml/ChowdhuryZW0C23} patch routing using a simplified MoE model, emphasizing class-discriminative patterns while reducing sample and model complexity.

% While these works offer a decompose learning paradigm, the method lacks prior applications to sentence embedding, necessitating a new design to reflect the linguistic nature of the problem. 
% %
% In particular, the method is inapplicable to our classify-and-rank setting due to the following issues.
%

While these works offer a decomposed learning paradigm, their approaches are inapplicable to the linguistic property of sentence relationship prediction. 
In particular, we must address the following issues to formulate a suitable space decomposition method. 
First, prior works lack an explicit control mechanism to differentiate between types of concern (upper-range and lower-range classes in this case). 
Second, the space embedding decomposition is only performed in the training step while aggregating the embedding space at the inference stage. 
In the following section, we propose a solution that addresses these limitations to reflect the problem~requirements.

%% file: 3_Methodology.tex
\section{Mixture of Specialized Projectors}
%
% Purpose of this paragraph:
% - แนะนำภาพรวมวิธีของเราโดยจะสรุปการออกแบบและส่วนประกอบของวิธีแก้ปัญหา
% - 
%
We design our method, \emph{Mixture of Specialized Projectors (MixSP)}, based on our linguistic observation that when labeling sentence pairs for similarity, the score ranges of [0,4) and [4,5] are considered as two distinct classes: lower-range and upper-range, respectively. 
In particular, we address our space decomposition problem by designing a classify-and-rank pipeline with a routing mechanism and one specialized projector for each class.
Consequently, we improve (i) the ability to differentiate between upper-range and lower-range sentence pairs and (ii) the ranking performance within each, thereby uplifting the overall sentence similarity prediction performance.

%In particular, we address our space decomposition problem by designing a classify-and-rank pipeline to address the issues raised in the previous section: (i) the lack of a mechanism to differentiate between positive and negative classes and (ii) the lack of specialized embedding space for each class.
%

Figure~\ref{fig:overview} displays our classify-and-rank pipeline consisting of the following components: 
\begin{compactitem}[\hspace{\setalign}•]
    \item The cross-encoder setup that transforms input sentence pairs into vectors (Section~\ref{subsec:cross-encoder}).
    \item The space decomposition mechanism that differentiates and separately handles upper-range and lower-range classes (Section \ref{subsec:space_decompose}).
    %\item Specialized projectors that separately handle negative and positive classes (Section \ref{subsec:expertise}). 
    \item The training objective that improves the ranking consistency within the embedding space of each class (Section \ref{subsec:representation learning}). 
\end{compactitem}

%
% Step 1: Accurately classify to reduce the overlap between the two classes
% Step 2: Construct a specialized embedding space for each class

\subsection{How Do We Encode Sentence Pairs?} \label{subsec:cross-encoder}
%
% Purpose of this paragraph:
% - อธิบายการ setup input ของ sentence pairs ก่อนที่จะโยนเข้า encoder
% -
%

% As shown in Figure~\ref{fig:overview}, our approach involves the development of a new MoE principle aimed at the complete separation of negative and positive sentence embedding spaces based on two key properties: (i) distinguishing the relationship between sent1 and sent2 (contradiction or entailment relations) and (ii) avoiding sum pooling to derive a soft representation, as in previous MoE studies.
%
% As shown in Figure~\ref{fig:overview}, our framework consists of 4 steps as follows: we first encode the sentence pairs in a cross-encoder paradigm with a pre-trained language model in Section \ref{subsec:cross-encoder}.
% %
% Then, the routing network separates representations into positive or negative groups, mitigating the collapse problem in the embedding space (Section \ref{subsec:routing}).
% %
% Expertise projects representations into positive or negative spaces, Section \ref{subsec:expertise}.
% %
% In Section \ref{subsec:representation learning}, we introduce an improved learning objective that improves the projected representations.

%
As shown in Figure~\ref{fig:overview}, given a sentence-pair ({\tt sent1},{\tt sent2}), we input them to a cross-encoder architecture from a pre-trained language model as {\tt [CLS]sent1[SEP]sent2[SEP]} and obtained three embeddings:
\begin{compactitem}[\hspace{\setalign}•]
    \item $\mathbf{h}_{\text{cls}}$ is the embedding of {\tt[CLS]} from the last layer of the pre-trained language model.
    \item $\mathbf{h}_{\text{x1}}$ is the mean pooling of {\tt sent1}'s representation.
    \item $\mathbf{h}_{\text{x2}}$ is the mean pooling of {\tt sent2}'s representation.
\end{compactitem}
%
% The main goal is to improve the ranking score between $\mathbf{h}_{\text{x1}}$ and $\mathbf{h}_{\text{x2}}$ while differentiate positive and negative classes. 

\subsection{How Do We Decompose the Embedding Space?} \label{subsec:space_decompose}

The embedding space decomposition mechanism consists of the routing network and specialized projectors.
These two parts are explained in the following subsections.

\subsubsection{The Routing Network} \label{subsec:routing}
%
% Purpose of this subsection:
% - อธิบาย routing network คืออะไร มีหน้าที่อะไร
% - อธิบายเหตุผลและรายละเอียดวิธีการ route ของเราว่ามันสามารถทำให้ไป projector แต่ละตัวได้ยังไง
% - รวมไปถึงการทำให้แต่ละคลาสสามารถไป projector ของคลาสตัวเองได้อย่างถูกต้อง
%

% As discussed in the related work section (Section~\ref{}), previous MoE works demonstrated the routing network's importance in separating representation into multiple groups to prevent the collapse problem in an embedding space.
%
As discussed in the related work section, the routing network can be used as a technique to decompose an embedding space.
Unlike existing methods, however, we introduce the {\tt [CLS]} label into the routing network in addition to the contextual representation $\mathbf{h}_{\text{x}}$.
This additional information allows the routing mechanism to understand the relation between the sentence pair ({\tt sent1},{\tt sent2}).
%
% However, the previous studies of the routing network used only the contextual representation $\mathbf{h}_{\text{x}}$, resulting in the routing network not understanding the relation between sent1 and sent2 (contradiction or entailment relations).
%
In particular, by incorporating the global representation, our sentence-pair input is the element-wise addictions: $\mathbf{h}_{\text{cls}}$ $\oplus$ $\mathbf{h}_{\text{x}}$.
We formulate the routing network as a group classification from a linear layer $\text{G}_1(\cdot)$, whether $\mathbf{h}_{\text{cls}} \oplus \mathbf{h}_{\text{x1}}$ and $\mathbf{h}_{\text{cls}} \oplus \mathbf{h}_{\text{x2}}$ are the representation of upper-range or lower-range groups based on the softmax probability $\hat{p}_{\text{x}j}$: 
\begin{equation} 
% \resizebox{1\hsize}{!}
\begin{aligned}
\hat{p}_{\text{x}j} = \text{SoftMax}(\text{G}_1(\mathbf{h}_{\text{cls}} \oplus \mathbf{h}_{\text{x}j})), 
\end{aligned}
\label{eq:softmax}
\end{equation}
where $j$ is {\tt sent1} ($j$=1) or {\tt sent2} ($j$=2).
We calculate the softmax probability $\hat{p}_{\text{x1}}$ and $\hat{p}_{\text{x2}}$ from {\tt sent1} $\mathbf{h}_{\text{x1}}$ and {\tt sent2} $\mathbf{h}_{\text{x2}}$, respectively.

To assist the routing network in classifying the input, we employ a binary cross-entropy (BCE) as follows.
%
%We also enforce the routing network to avoid one-node prediction over the others (the routing network always predicts the same node for every representation).
%
%We employ a binary cross-entropy (BCE) as a node balancing loss as follows:
%
\begin{equation} 
% \resizebox{1\hsize}{!}
\begin{aligned}
\mathcal{L}_{\text{Clf}}= \frac{1}{2} \text{BCE}(\hat{y}, \hat{p}_\text{x1}) + \frac{1}{2} \text{BCE}(\hat{y}, \hat{p}_\text{x2}),
\end{aligned}
\label{eq:RN_loss}
\end{equation}
where $y$ is a gold label indicating whether the sentence pair is upper-range or lower-range. 
%
% We minimize the discrepancy between the gold label $y$ and the softmax prediction from the routing network $\hat{p}_{\text{x}j}$.
%
The classification loss $\mathcal{L}_{\text{Clf}}$ is used as part of the overall learning objective explained in Section~\ref{subsec:representation learning}.

\subsubsection{Specialized Projectors} \label{subsec:expertise}
%
% Purpose of this subsection:
% - อธิบาย specialized projector คืออะไร มีหน้าที่ทำอะไร รายละเอียดการทำงานยังไง
% - รวมไปถึงเหตุผลว่าวิธีการของเราทำไมต้องแตกต่างจากวิธีอื่น
%

%
The main goal of MixSP is to decompose the representation $\mathbf{h}_{\text{x}j}$ into upper-range or lower-range subspaces. 
Note that previous works in embedding space decomposition produce a composite representation, i.e., obtaining the final representation by computing the vector summation from multiple projectors' outputs.
We found that such a soft-selection approach results in overlaps between subspaces, which is detrimental to the model's performance.
Consequently, we derive a hard-selection process in which our specialized projectors have a separate projection for each class and use only one head per representation.
%
%In particular, we categorize the representations $\mathbf{h}_{\text{x1}}$ and $\mathbf{h}_{\text{x2}}$ to specialized projectors following the output from the routing network $\hat{p}_{\text{x1}}$ and $\hat{p}_{\text{x2}}$, respectively.
%
% In particular, the specialized projectors have a separate project for each class and use only one head per representation (hard selection).
% %
% On the other hand, existing works produced the final representation by combining multiple projectors' outputs and obtaining a composite representation from them (soft selection).
%
% In our investigation, we found that such the overlap of hard selection can be detrimental to the model's performance.
%
%In contrast, previous works simply formulated the final representation by summing multiple projectors' output (sum pooling) and obtained a soft representation (every projector has a participant in the final representation). 
%
% This can lead us to a poor performance since the composite representation. 
%This presents a dataflow challenge since subspace overlap hurts the performance.
%
% In contrast, we design a new MoE data flow for sentence embedding 
%
%
In particular, our two specialized projectors, (i) an upper-range projector $\text{Upper}(\cdot)$ and (ii) a lower-range projector $\text{Lower}(\cdot)$, map representations $\mathbf{h}_{\text{x}j}$ to upper-range or lower-range subspaces as follows: 
%
% \emph{We can see that our specialized projectors are a hard version of decomposition techniques} since we split the positive and negative representations completely. 
%
% We define the projected representation of $\text{E}(\cdot)$ as follows:
%
% \begin{equation} 
% \centering
% % \resizebox{1\hsize}{!}
% \begin{aligned}
% \mathbf{z}_{\text{x}j} = \text{E}_{\text{k}}(\mathbf{h}_{\text{x}j})*\beta_{\text{j}},
% \end{aligned}
% \label{eq:expertise}
% \end{equation}
%
\begin{equation}
\mathbf{z}_{\text{x}j}= \begin{cases} \text{Upper}(\mathbf{h}_{\text{x}j})*\beta_{\text{j}}, & \text { if } \text{argmax}(\hat{p}_{\text{x}j}) = 0 \\ \text{Lower}(\mathbf{h}_{\text{x}j})*\beta_{\text{j}}, & \text { otherwise }\end{cases}
\label{eq:expertise}
\end{equation}
where $\beta_j$ is the highest probability of $\hat{p}$ obtained from max($\hat{p}_{\text{x}j}$) and $\mathbf{z}_{\text{x}j}$ is the representation that mapped to upper-range or lower-range subspaces.

With the output from different projectors, we obtain the representation pair of upper-range and lower-range samples separately, $\mathbf{z}_{\text{x1}}$ and $\mathbf{z}_{\text{x2}}$.
However, $\mathbf{z}_{\text{x}j}$ is produced from a random weight of the specialized projectors.
We require a method to improve the representation of the projectors.

\subsection{How Do We Improve The Contextual Embedding Space?} \label{subsec:representation learning}
%
% Purpose of this subsection:
% - อธิบาย learning objective คืออะไร มีหน้าที่ทำอะไร
% - รวมไปถึงเหตุผลว่าวิธีการของเราทำไมต้องแตกต่างจากวิธีอื่น
%

One of the key components in this work is improving the semantic understanding of representation $\mathbf{z}_{\text{x}j}$ produced from specialized projectors $\text{Upper}(\cdot)$ or $\text{Lower}(\cdot)$.
A common practice is applying supervised contrastive learning to a pair-wise representation.
However, we found that in-batch contrastive learning harms the projectors' performance because it is required to compose the representations from difference projectors ($\text{Upper}(\cdot)$ and $\text{Lower}(\cdot)$) in the same mini-batch.
(See Section~\ref{subsec:ablation_study} for experimental analysis.)
%
% The representations from different projectors are composted with contrastive learning and 
%
Therefore, we design a more suitable learning objective for the classify-and-rank mechanism, which is linear similarity prediction for each projector separately.
In particular, we concatenate $\mathbf{z}_{\text{x1}}$ with $\mathbf{z}_{\text{x2}}$ using a linear layer $\text{G}_2$ where the linear's output is regression number from zero (dissimilar) to one (similar).
We then minimize the discrepancy between the output and gold label $y_\text{sim}$ (STS score) with the BCE loss as follows:
\begin{equation} 
% \resizebox{1\hsize}{!}
\begin{aligned}
\mathcal{L}_{\mathrm{RL}}= \text{BCE}(y_\text{sim}, \text{G}_2(\text{concat}(\mathbf{z}_{\text{x1}},\mathbf{z}_{\text{x2}})))
\end{aligned}
\label{eq:RL_loss}
\end{equation}
The final training loss $\mathcal{L}$ is an end-to-end paradigm of representation learning and classification losses:
\begin{equation} 
% \resizebox{1\hsize}{!}
\begin{aligned}
\mathcal{L} = \underbrace{\alpha_1 \mathcal{L}_{\mathrm{RL}}}_{\text{representation learning}} + \underbrace{\alpha_2 \mathcal{L}_{\text{Clf}}}_{\text{classification}}
\end{aligned}
\label{eq:MixSE_loss}
\end{equation}
The parameters $\alpha_1$ and $\alpha_2$ are the loss weights obtained from tuning on the development set.

%% file: 4_Exp_setup.tex
\section{Experimental Setup}

The purpose of our experimental studies is to understand how MixSP performs compared to the traditional fine-tuning method~\cite{reimers-2019-sentence-bert} and Mixture-of-Expert~\cite{NEURIPS2022_2f00ecd7} as competitive methods.
Since we present MixSP as a generic STS enhancement method, we assess MixSP against its competitive methods by varying critical factors like the pre-trained sentence encoder, base model, and evaluation tasks and observe how results generalize. 
%
% In addition, the experimental studies also include an in-depth component analysis to provide insight into our design decisions.

\subsection{Competitive Methods} \label{subsec:comp}

To assess the effectiveness of MixSP as an STS enhancement method, we compare it against two competitors.
Note that for the full implementation of competitive methods, please refer to Appendix~\ref{appendix:competitive_methods}.
\begin{compactitem}[\hspace{\setalign}•]
    \item \textbf{+FT}~\cite{reimers-2019-sentence-bert} [No Space Decomposition]. 
    We fine-tune the base model through the cosine similarity function. 
    This method serves as our fine-tuning baseline in which the base model is directly adapted to the STS task.
    
    \item \textbf{+MoE}~\cite{NEURIPS2022_2f00ecd7} [Soft Selection]. 
    As our comparator for embedding space decomposition, we employ the Mixture-of-Expert method.
    Its relative performance against MixSP will provide insight into the merit of the hard selection approach adopted by MixSP, i.e., supervised classification loss $\mathcal{L}_{\text{Clf}}$ and the Argmax selection as opposed to weighted-average pooling in Section~\ref{subsec:expertise}.
    %
    % Additionally, the routing network includes a supervised signal to enhance the model's understanding of sentence relationships.
    %
\end{compactitem}
To ensure fair and transparent assessment, we apply the same STS dataset, STS-B~\cite{cer-etal-2017-semeval}, to all methods.
We chose STS-B due to its well-documented sources, which can help us avoid data leakage when selecting evaluation datasets. 
For the full data leakage discussion, please refer to Appendix~\ref{appendix:data_leakage}.
 
%
%STS-B is a collection of STS samples compiled from STS 2021-2016; as a result, any test sets overlapping with these datasets are excluded from our assessment. 

%With these fine-tuning competitive methods, we can ensure that all methods have the same base model and data while we only vary the fine-tuning methods in our experimental results. 

\subsection{Training Setup}
We use STS-B training data following prior works~\cite{cer-etal-2017-semeval, reimers-2019-sentence-bert}. 
For the lower-range and upper-range samples, we separate the lower-range and upper-range samples according to the STS score in the range of [0,4) and [4,5], respectively. 
We use AamW as the optimizer, a learning rate of 5$e^{-5}$, and a batch size of 16 for 10 epochs.
We use $\alpha_1$ and $\alpha_2$ equal to 7$e^{-4}$ and 1$e^{-4}$, respectively (tuned on the STS-B development set). 
For the base encoder, we use SBERT, SimCSE, and DiffAug to observe the improvement of changing from a single embedding to a separate embedding space.
All experiments were done on a single V100 with three random seeds per model.

\subsection{Sentence Encoders and Base Models} 
%
% Purpose of this paragraph:
% - อธิบายถึงคู่แข่งที่จะเอาเปรียบเทียบรวมไปถึงรายละเอียดเทคนิควิธีการและ dataset ที่คู่แข่งใช้

We employ off-the-shelf text encoder models as follows:
\begin{compactitem}[\hspace{\setalign}•]
    \item \textbf{SBERT}~\cite{reimers-2019-sentence-bert}. A supervised baseline. The model was trained on the STS-B training set with cosine similarity as the training objective (similar to our method).
    \item \textbf{SimCSE}~\cite{DBLP:conf/emnlp/GaoYC21}. A simple contrastive learning method using dropout as the data augmentation.
    \item \textbf{DiffAug}~\cite{wang2022diffaug}. A two-stage contrastive learning framework. Contrastive learning is applied to minimize the discrepancy between two differentiable augmentation schemes. 
\end{compactitem}
Note that SimCSE and DiffAug were trained on NLI-supervised datasets, MNLI~\cite{williams-etal-2018-broad} and SNLI~\cite{bowman-etal-2015-large} datasets.
In addition to varying the sentence encoders, we test our method with two different architectures, BERT-Base and RoBERTa-Base. 

\subsection{Evaluation Tasks} \label{subsec:eval_setup}
%
% Purpose of this subsection:
% - อธิบายวิธีการ evaluate ในงานของเราและใช้ dataset อะไร
% - รวมไปถึง task อื่นๆ เพื่อทำให้งานเรามีความน่าเชื่อถือมากขึ้น
%

We evaluate the effectiveness of our method compared to competitive methods on two tasks: STS and zero-shot tasks.
For the STS task, we select the STS benchmarks with low to non-word overlapping between our training data and benchmarks to prevent data leakage, as discussed in Appendix~\ref{appendix:data_leakage}. 
In particular, we evaluate our model with three STS benchmarks: CDSC-R (validation set), CDSC-R (test set)~\cite{wroblewska-krasnowska-kieras-2017-polish}, and BIOSSES~\cite{10.1093/bioinformatics/btx238}.
In addition, we also evaluate our model on standard seven STS datasets in Appendix~\ref{subsec:sts_standard}.
We use Spearman's rank correlation as the main metric to be consistent with prior works.

For the zero-shot task, we assess the generalizability of our model across unseen tasks/domains, namely reranking and binary text classification.
In the reranking task, we adopt the settings and datasets from MTEB~\cite{muennighoff-etal-2023-mteb}, where the Mean Average Precision (MAP) is the main evaluation metric.
We also test our model on sentence-pair binary classification tasks where the model has to decide if a sentence-pair has certain relations.
The ground truth labels of this task are either 0 or 1. 
We calculate Area Under Curve (AUC) scores with the binary labels and the relevance scores predicted by models following previous works~\cite{DBLP:conf/emnlp/LiZHWYL20, DBLP:conf/iclr/0001JMYH22,limkonchotiwat-etal-2023-sct}.

\begin{table*}[ht]
\hspace*{-4mm}
\centering
\setlength\doublerulesep{3pt}
\scalebox{0.80}{
\setlength{\tabcolsep}{8pt}
%\fontsize{10pt}{12pt}\selectfont
\begin{tabular}{c|c|c|c|c|c|c|c|c}
\hline
\multirow{3}{*}{\textbf{Method}} &
  \multicolumn{4}{c|}{\textbf{BERT$\textbf{-}$Base}} &
  \multicolumn{4}{c}{\textbf{RoBERTa$\textbf{-}$Base}} \\
 \cline{2-9}& \multirow{2}{*}{\textbf{BIOSSES} }& 
 \textbf{CDSC$\textbf{-}$R} & 
 \textbf{CDSC$\textbf{-}$R} & 
 \multirow{2}{*}{\textbf{Avg.}} & 
 \multirow{2}{*}{\textbf{BIOSSES}} & 
 \textbf{CDSC$\textbf{-}$R} &
 \textbf{CDSC$\textbf{-}$R} &
 \multirow{2}{*}{\textbf{Avg.}} \\ &
  \multicolumn{1}{c|}{} &
  \multicolumn{1}{c|}{\textbf{(Val)}} &
  \multicolumn{1}{c|}{\textbf{(Test)}} &
  &
  \multicolumn{1}{c|}{} &
  \multicolumn{1}{c|}{\textbf{(Val)}} &
  \multicolumn{1}{c|}{\textbf{(Test)}} &
  \\ \hline \hline
\multicolumn{9}{c}{\textit{SBERT as the base encoder}} \\ \hline
\multicolumn{1}{l|}{SBERT} &
  \multicolumn{1}{c|}{63.88} &
  \multicolumn{1}{c|}{59.48} &
  \multicolumn{1}{c|}{63.53} &
  \multicolumn{1}{c|}{62.30} &
  \multicolumn{1}{c|}{72.03} &
  \multicolumn{1}{c|}{68.37} &
  \multicolumn{1}{c|}{70.57} &
  70.32 \\ 
\multicolumn{1}{l|}{$\textbf{+}$MoE} &
  \multicolumn{1}{c|}{78.52} &
  \multicolumn{1}{c|}{84.70} &
  \multicolumn{1}{c|}{84.02} &
  \multicolumn{1}{c|}{82.41} &
  \multicolumn{1}{c|}{73.75} &
  \multicolumn{1}{c|}{85.09} &
  \multicolumn{1}{c|}{79.88} &
  79.57 \\ 
\multicolumn{1}{l|}{$\textbf{+}$MixSP} &
  \multicolumn{1}{c|}{\textbf{80.58$_{\pm0.64}$}} &
  \multicolumn{1}{c|}{\textbf{85.08$_{\pm0.65}$}} &
  \multicolumn{1}{c|}{\textbf{84.15$_{\pm0.59}$}} &
  \multicolumn{1}{c|}{\textbf{83.27$_{\pm0.58}$}} &
  \multicolumn{1}{c|}{\textbf{76.01$_{\pm1.13}$}} &
  \multicolumn{1}{c|}{\textbf{85.60$_{\pm0.44}$}} &
  \multicolumn{1}{c|}{\textbf{81.21$_{\pm0.30}$}} &
  \multicolumn{1}{c}{\textbf{80.94$_{\pm0.47}$}} \\ \hline
\multicolumn{9}{c}{\textit{SimCSE as the base encoder}} \\ \hline
\multicolumn{1}{l|}{SimCSE} &
  \multicolumn{1}{c|}{68.38} &
  \multicolumn{1}{c|}{70.21} &
  \multicolumn{1}{c|}{70.63} &
  \multicolumn{1}{c|}{69.74} &
  \multicolumn{1}{c|}{67.75} &
  \multicolumn{1}{c|}{68.38} &
  \multicolumn{1}{c|}{70.64} &
  68.92 \\ 
\multicolumn{1}{l|}{$\textbf{+}$FT} &
  \multicolumn{1}{c|}{76.62} &
  \multicolumn{1}{c|}{69.98} &
  \multicolumn{1}{c|}{69.53} &
  \multicolumn{1}{c|}{72.04} &
  \multicolumn{1}{c|}{73.35} &
  \multicolumn{1}{c|}{69.01} &
  \multicolumn{1}{c|}{71.69} &
  71.35 \\ 
\multicolumn{1}{l|}{$\textbf{+}$MoE} &
  \multicolumn{1}{c|}{77.07} &
  \multicolumn{1}{c|}{82.87} &
  \multicolumn{1}{c|}{83.34} &
  \multicolumn{1}{c|}{81.09} &
  \multicolumn{1}{c|}{72.65} &
  \multicolumn{1}{c|}{84.01} &
  \multicolumn{1}{c|}{80.33} &
  79.00 \\ 
\multicolumn{1}{l|}{$\textbf{+}$MixSP} &
  \multicolumn{1}{c|}{\textbf{82.61$_{\pm0.80}$}} &
  \multicolumn{1}{c|}{\textbf{88.27$_{\pm0.48}$}} &
  \multicolumn{1}{c|}{\textbf{85.28$_{\pm0.06}$}} &
  \multicolumn{1}{c|}{\textbf{85.39$_{\pm0.35}$}} &
  \multicolumn{1}{c|}{\textbf{80.74$_{\pm0.85}$}} &
  \multicolumn{1}{c|}{\textbf{84.48$_{\pm0.31}$}} &
  \multicolumn{1}{c|}{\textbf{80.41$_{\pm0.19}$}} &
  \multicolumn{1}{c}{\textbf{81.88$_{\pm0.34}$}} \\ \hline
\multicolumn{9}{c}{\textit{DiffAug as the base encoder}} \\ \hline
\multicolumn{1}{l|}{DiffAug} &
  \multicolumn{1}{c|}{40.12} &
  \multicolumn{1}{c|}{61.42} &
  \multicolumn{1}{c|}{62.61} &
  \multicolumn{1}{c|}{54.72} &
  \multicolumn{1}{c|}{39.15} &
  \multicolumn{1}{c|}{62.47} &
  \multicolumn{1}{c|}{64.65} &
  55.42 \\ 
\multicolumn{1}{l|}{$\textbf{+}$FT} &
  \multicolumn{1}{c|}{71.26} &
  \multicolumn{1}{c|}{67.91} &
  \multicolumn{1}{c|}{70.25} &
  \multicolumn{1}{c|}{69.81} &
  \multicolumn{1}{c|}{71.02} &
  \multicolumn{1}{c|}{64.14} &
  \multicolumn{1}{c|}{70.66} &
  68.61 \\ 
\multicolumn{1}{l|}{$\textbf{+}$MoE} &
  \multicolumn{1}{c|}{79.95} &
  \multicolumn{1}{c|}{86.29} &
  \multicolumn{1}{c|}{84.71} &
  \multicolumn{1}{c|}{83.65} &
  \multicolumn{1}{c|}{72.65} &
  \multicolumn{1}{c|}{85.01} &
  \multicolumn{1}{c|}{81.33} &
  79.66 \\ 
\multicolumn{1}{l|}{$\textbf{+}$MixSP} &
  \multicolumn{1}{c|}{\textbf{81.23$_{\pm0.84}$}} &
  \multicolumn{1}{c|}{\textbf{88.28$_{\pm0.42}$}} &  
  \multicolumn{1}{c|}{\textbf{85.45$_{\pm0.46}$}} &
  \multicolumn{1}{c|}{\textbf{84.99$_{\pm1.03}$}} &
  \multicolumn{1}{c|}{\textbf{80.35$_{\pm0.65}$}} &
  \multicolumn{1}{c|}{\textbf{86.16$_{\pm0.58}$}} &
  \multicolumn{1}{c|}{\textbf{81.79$_{\pm0.83}$}} &
  \multicolumn{1}{c}{\textbf{82.77$_{\pm0.69}$}} \\ \hline \hline
\end{tabular}
}
\vspace{-2mm}
\caption{Spearman's rank correlation on the STS benchmarks in a fair environment for assessment.}
\vspace{-4mm}
\label{tab:sts}
\end{table*}

%% file: 5_Exp_results.tex
\section{Experimental Results}

\subsection{STS Benchmarks}
%
% Purpose of this paragraph:
% - อธิบายภาพรวมผลลัพธ์การทดลองของเรา
% - รวมไปถึงอธิบายเหตุผลที่คะแนนเพิ่มขึ้นไม่ใช่เพราะ dataset แต่เป็นเพราะการออกแบบ method
%

%
% In this study, we evaluate the effectiveness of our method compared to competitive methods on seven STS benchmarks.
%
% As shown in Table~\ref{tab:sts}, MixSP outperforms competitive methods in the average score.
%
%
%These results support MixSP as a generic STS enhancement method.
%

To assess the generalizability of MixSP, we tested it on two architectures, three base encoders, and three datasets, bringing the total number of combinations to 18.
%
% As shown in Table~\ref{tab:sts}, MixSP outperforms competitive methods in 17 out of 18 cases.
%
As shown in Table~\ref{tab:sts}, MixSP outperforms competitive methods in all cases.
%
%The combination SimCSE+MixSP is the best performer for BERT-Base, and DiffAug+MixSP is the best performer for RoBERTa-Base, averaging the three datasets' results. 
%
%We can see that while FT provides performance improvement in 16 out of 18 cases, the method is outperformed by MoE in 17 out of 18 cases and MixSP in all cases.
%
%These results demonstrate the effectiveness of the two mixture-based methods. 
%
%For SimCSE, the performance gap between MixSP and MoE is 4.30 points for BERT-Base and 2.88 points for RoBERTa-Base, averaging the three datasets' results.
%
Moreover, as a generic improvement method, we can observe reasonable improvement compared to fine-tuning techniques for all the average scores. 
For instance, when changing from a single to separate embedding spaces, we improved the performance of SOTA (SimCSE-BERT-Base) from 69.74 to 85.39 (15.65 improvement points).
In particular, compared to fine-tuning methods that use the same dataset as our method, FT and MoE, we outperform them by 13.35 points and 4.30 points on the average score, respectively. 
Furthermore, we evaluate the effectiveness of each method on RoBERTa-Base.
The experimental results demonstrate consistency improvement similar to BERT-Base, e.g., MixSP outperforms FT and MoE on DiffAug by 14.16 and 3.11 points on the average score, respectively. 
%
% \emph{These results emphasize that MixSP outperforms competitive methods because we recast the sentence embedding problem to a smarter task.}
%
% In addition, our method is insensitive to all baseline encoders and generalization to out-of-domain data.  
%
Note that we experimented on the accuracy of our router network in Appendix~\ref{subsubsec:routing}.
%
% Additionally, we expanded our evaluation of each model's effectiveness on the seven STS benchmark shown in Table~\ref{tab:base_encoder}.

% The results presented in Table~\ref{tab:base_encoder} consistently demonstrate the effectiveness of our approach across various base encoders similar to BERT-Base. 

%
% Note that sick-r's lower performance be attributed to dataset differences, similar with NLI and thus resulting in higher scores for methods using the NLI dataset, such as SimCSE, PromptBERT, and DiffAug, compared to our method.
%
% Note that the lower performance on SICK-R compared to contrastive methods can be attributed to the training data of contrastive methods being similar to SICK-R (See Appendix~\ref{appendix:data_choice} for data choice analysis).
%
% Note that SICK-R differs from other datasets regarding the STS score annotation process. 
%
% In particular, their sentence pairs were originally labeled for NLI. 

\subsection{Zero-shot Downstream Tasks}
%
% Purpose of this paragraph:
% - ต้องการที่จะสื่อว่างานเราคะแนนไม่ได้ดีขึ้นแค่ in-domain (sts score) แต่มันยัง generalize กว่าโมเดลคู่แข่ง
% - 
%
In our prior experiment, we demonstrated the effectiveness of our method in the seen task (STS). 
%
% However, a crucial question emerges: \emph{does sentence embedding performance in the STS task accurately represent a model's capabilities?} 
However, a crucial question emerges: \emph{does sentence embedding performance in the STS task accurately represent a model's capabilities?} 
%
% Moreover, whether our specialized project, MixSP, will perform better or worse on the tasks that do not have a continuous scale of 0 to 5.
%
To explore this, we assess our method in zero-shot settings, evaluating its performance in two unseen tasks/domains: \emph{reranking} and \emph{binary text classification}. 
This study aims to unveil the versatility of our model, offering insights into its potential application across a diverse range of tasks and domains.

\subsubsection{Reranking}

In this experiment, we study the effectiveness of our model on unseen datasets and tasks from the MTEB reranking benchmark.
%
% The goal is to prioritize and rank the results based on their relevance to the query.
% %
% Our method classified each query and text, whether the text pair was upper-range or lower-range (relevant or irrelevant), and then produced a similarity score between zero and one for each pair.
% %
% We ranked them according to the score (from high to low) to obtain the final result. 
%
As demonstrated in Table~\ref{tab:reranking}, MixSP outperforms competitive methods in all cases.
Our method improves the performance of SOTA (SimCSE) from 47.54 to 51.01 points.
Moreover, we achieve a new SOTA on DiffAug by improving the performance from 47.38 to 52.94 points.
In contrast, we found performance decreases in the traditional fine-tuning technique, FT. 
For example, when we applied FT to DiffAug, the performance of DiffAug decreased from 47.38 to 46.66.
%
% These results emphasize that our \textit{classify-and-rank} mechanism improves the robustness of the ranking task by separating irrelevant from relevant samples while the ranking of relevant samples is consistently higher than the other.  
%
% \emph{This finding emphasizes a noteworthy contradiction in previous works, emphasizing that increased complexity does not necessarily lead to improved performance.}
%
\emph{This finding demonstrates that MixSP has benefits beyond the seen task of STS.}

%
% \emph{This finding emphasizes that only improving the performance on STS did not reflect other downstream tasks' performance}.
% %
% In contrast, MixSP consistently improves the performance on all tasks.

% \begin{table}[h]

% \hspace*{-4mm}
% \centering
% \setlength\doublerulesep{3pt}
% \scalebox{0.80}{
% \setlength{\tabcolsep}{4pt}
% %\fontsize{10pt}{12pt}\selectfont
% \begin{tabular}{l|c|c|c|c|c}
% \hline
% \textbf{Method} & \textbf{AU} & \textbf{MM} & \textbf{SD} & \textbf{SO} & \textbf{Avg.} \\ \hline \hline
% SBERT           & 51.09       & 30.24       & 69.40       & 36.54       & 46.82         \\ 
% SimCSE          & 51.80       & 29.30       & 70.14       & 38.90       & 47.54         \\ 
% PromptBERT      & 52.22       & 29.28       & 70.13       & 38.48       & 47.53         \\ 
% DiffAug         & 51.10       & 29.52       & 71.10       & 37.81       & 47.38         \\ 
% MixSP (ours)     & \textbf{53.64}  & \textbf{30.49}  & \textbf{76.05}  & \textbf{41.69}  & \textbf{50.47} \\ \hline \hline
% \end{tabular}}
% \vspace{-2mm}
% \caption{ The MAP score on the reranking task from the MTEB benchmark where AU = AskUbuntuDupQuestions, MM = MindSmallReranking, SD = SciDocsRR, and SO = StackOverflowDupQuestions.}
% \vspace{-5mm}
% \label{tab:reranking}
% \end{table}

\begin{table}[h]
\hspace*{-4mm}
\centering
\setlength\doublerulesep{3pt}
\scalebox{0.80}{
\setlength{\tabcolsep}{4pt}
%\fontsize{10pt}{12pt}\selectfont
\begin{tabular}{l|c|c|c|c|c}
\hline
\textbf{Method} & \textbf{AU} & \textbf{MM} & \textbf{SD} & \textbf{SO} & \textbf{Avg.} \\ \hline \hline
\multicolumn{6}{l}{\textit{SBERT-BERT-Base}} \\ \hline
SBERT           & 51.09       & 30.24       & 69.40       & 36.54       & 46.82         \\ 
+MoE            & 53.72       & 30.00       & 75.14       & 41.72       & 50.15         \\
+MixSP          & \textbf{54.56} & \textbf{30.59} & \textbf{75.78} & \textbf{42.39} & \textbf{50.83}  \\
\hline
\multicolumn{6}{l}{\textit{SimCSE-BERT-Base}} \\ \hline
SimCSE          & 51.80       & 29.30       & 70.14       & 38.90       & 47.54         \\
+FT             & 51.81       & 28.91       & 70.07       & 38.19       & 47.25         \\
+MoE            & 53.94       & 27.98       & 74.25       & 41.51       & 49.42         \\
+MixSP          & \textbf{54.50}       & \textbf{30.68}       & \textbf{75.35}       & \textbf{43.51}       & \textbf{51.01}         \\
\hline
\multicolumn{6}{l}{\textit{DiffAug-BERT-Base}} \\ \hline
DiffAug         & 51.10       & 29.52       & 71.10       & 37.81       & 47.38         \\ 
+FT             & 51.04       & 29.44       & 69.82       & 36.34       & 46.66         \\
+MoE            & 52.68       & 29.83       & 74.76       & 40.52       & 49.45         \\
+MixSP          & \textbf{53.92}       & \textbf{30.07}       & \textbf{75.68}       & \textbf{42.08}       & \textbf{52.94}         \\
\hline \hline
\end{tabular}}
\vspace{-2mm}
\caption{ The MAP score on the reranking task from the MTEB benchmark where AU = AskUbuntuDupQuestions, MM = MindSmallReranking, SD = SciDocsRR, and SO = StackOverflowDupQuestions.}
\vspace{-5mm}
\label{tab:reranking}
\end{table}

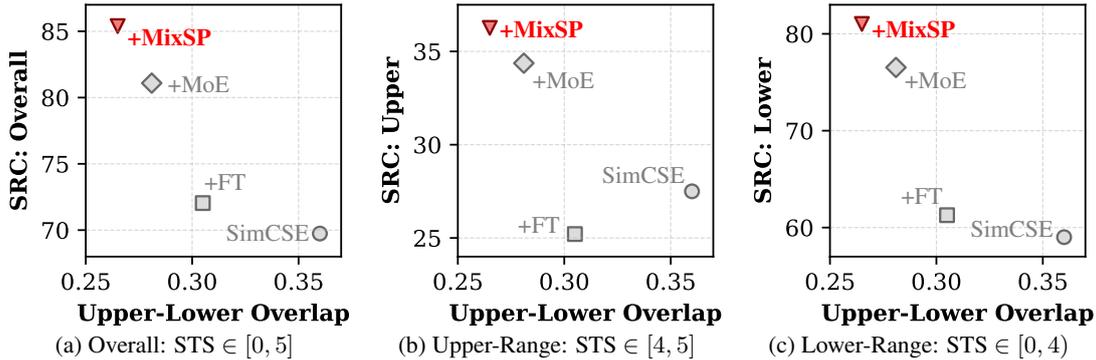
\begin{figure*}[ht]
\hspace*{-4mm}
\centering
\newcommand{\subfigwidth}{0.30\textwidth}
\newcommand{\legendfontsize}{\footnotesize}

\begin{subfigure}{\subfigwidth}
\vspace{-4mm}
\centering
      \begin{overpic}[width=\textwidth]{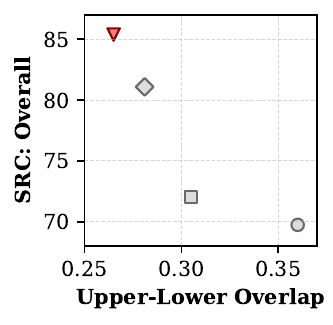}
        \put(37,82){\bf \color{red} \legendfontsize +MixSP}
        \put(48,69){\color{gray} \legendfontsize +MoE}
        \put(58,42){\color{gray} \legendfontsize +FT}
        \put(64,28){\color{gray} \legendfontsize SimCSE}
      \end{overpic}
\vspace{-8mm}
\caption{Overall: STS $\in [0,5]$}
\label{fig:overall_overlapped}
\end{subfigure}
\begin{subfigure}{\subfigwidth}
\vspace{1mm}
\centering
      \begin{overpic}[width=\textwidth]{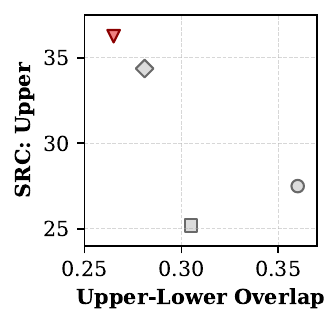}
        \put(37,84){\bf \color{red} \legendfontsize +MixSP}
        \put(46,70){\color{gray} \legendfontsize +MoE}
        \put(42,30){\color{gray} \legendfontsize +FT}
        \put(65,44){\color{gray} \legendfontsize SimCSE}
      \end{overpic}
\vspace{-8mm}
\caption{Upper-Range: STS $\in [4,5]$}
\label{fig:pos}
\end{subfigure}
\begin{subfigure}{\subfigwidth}
\vspace{1mm}
\centering
      \begin{overpic}[width=\textwidth]{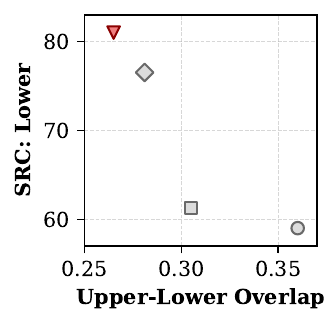}
        \put(37,84){\bf \color{red} \legendfontsize +MixSP}
        \put(46,70){\color{gray} \legendfontsize +MoE}
        \put(45,38){\color{gray} \legendfontsize +FT}
        \put(64,29){\color{gray} \legendfontsize SimCSE}
      \end{overpic}
\vspace{-8mm}
\caption{Lower-Range: STS $\in [0,4)$}
\label{fig:neg}
\end{subfigure}
\vspace{-3mm}
\caption{Comparison between MixSP and competitors on three STS datasets. 
The x-axis represents class overlap, quantifying the confusion between upper-range and lower-range classes. 
The y-axis measures similarity ranking performance using Spearman's rank correlation (SRC) coefficients. 
The figures include SRC Coefficients for Overall Results (\ref{fig:overall_overlapped}), Upper-Range (\ref{fig:pos}), and Lower-Range (\ref{fig:neg}). These visuals illustrate the reduced class overlap and effective sentence pair ranking compared to competitors.}
\vspace{-2mm}
\label{fig:overlap}
\end{figure*}

\subsubsection{Binary Text Classification}
%
% Purpose of this paragraph:
% - อธิบายผลการทดลองใน binary text classification task 
% - รวมไปถึงทำแล้วได้อะไร
%

%
In this experiment, we study the generalization of sentence embedding methods on standard binary text classification datasets.
%
% We conducted a study similar to the ranking task, and MixSP assigned the similarity score of each sentence pair.
%
As shown in Table~\ref{tab:binary}, our method outperforms competitive methods in the average score.
In particular, we improved the average AUC score of DiffAug-FT and DiffAug-MoE by 4.18 and 0.58 points, respectively.
Moreover, a consistent pattern emerges, similar to observations in the STS and reranking benchmarks.
Our method consistently outperforms competitive methods on SBERT and SimCSE; e.g., the gap between our work and MoE on SBERT is 0.68 points.
%
% Thus, we can answer the questions at the beginning of the zero-shot experiment that \emph{our method exhibits superior generalization to both seen and unseen domains/tasks when compared to competitive alternatives}.

% \begin{table}[h]
% \vspace{-1mm}
% \hspace*{-4mm}
% \centering
% \setlength\doublerulesep{3pt}
% \scalebox{0.80}{
% \setlength{\tabcolsep}{5pt}
% %\fontsize{10pt}{12pt}\selectfont
% \begin{tabular}{l|c|c|c|c}
% \hline
% \textbf{Method} & \multicolumn{1}{c|}{\textbf{QQP}} & \multicolumn{1}{c|}{\textbf{QNLI}} & \multicolumn{1}{c|}{\textbf{MRPC}} & \multicolumn{1}{c}{\textbf{Avg.}} \\ \hline \hline
% SBERT       & 80.44          & 68.51          & 80.13          & 76.36          \\ 
% SimCSE      & 81.37          & 74.53          & 77.83          & 77.91          \\ 
% PromptBERT  & 81.49          & 75.22          & 77.31          & 78.01          \\ 
% DiffAug     & \textbf{81.90}          & 74.17          & 78.08          & 78.05          \\ 
% MixSP (ours) & 81.30 & \textbf{79.75} & \textbf{82.75} & \textbf{81.27} \\ \hline \hline
% \end{tabular}
% }
% \vspace{-2mm}
% \caption{ The AUC score of binary text classification on three standard datasets.}
% \vspace{-5mm}
% \label{tab:binary}
% \end{table}

\begin{table}[h]
% \vspace{-1mm}
\hspace*{-4mm}
\centering
\setlength\doublerulesep{3pt}
\scalebox{0.80}{
\setlength{\tabcolsep}{5pt}
%\fontsize{10pt}{12pt}\selectfont
\begin{tabular}{l|c|c|c|c}
\hline
\textbf{Method} & \multicolumn{1}{c|}{\textbf{QQP}} & \multicolumn{1}{c|}{\textbf{QNLI}} & \multicolumn{1}{c|}{\textbf{MRPC}} & \multicolumn{1}{c}{\textbf{Avg.}} \\ \hline \hline
\multicolumn{5}{l}{\textit{SBERT-BERT-Base}} \\ \hline
SBERT     & 80.44          & 68.51          & 80.13          & 76.36       \\ 
+MoE      & 81.60          & 76.91          & 83.08          & 80.53       \\ 
+MixSP    & \textbf{82.29}          & \textbf{78.21}          & \textbf{83.21}          & \textbf{81.24}       \\ \hline
\multicolumn{5}{l}{\textit{SimCSE-BERT-Base}} \\ \hline
SimCSE    & 81.37          & 74.53          & 77.83          & 77.91       \\
+FT       & \textbf{82.55}          & 72.40          & 81.48          & 78.81       \\ 
+MoE    & 81.66          & 79.36          & \textbf{82.54}          & 81.19       \\
+MixSP          & 82.08          & \textbf{80.14}          & 82.32          & \textbf{81.51}       \\ \hline
\multicolumn{5}{l}{\textit{DiffAug-BERT-Base}} \\ \hline
DiffAug   & 81.90          & 74.17          & 78.08          & 78.05       \\
+FT       & 81.39          & 69.11          & 81.30          & 77.27       \\ 
+MoE    & 81.33          & 78.64          & \textbf{82.63}          & 80.87       \\
+MixSP          & \textbf{81.85}          & \textbf{80.11}          & 82.40          & \textbf{81.45}       \\ \hline
\hline \hline
\end{tabular}
}
\vspace{-2mm}
\caption{ The AUC score of binary text classification on three standard datasets.}
\vspace{-5mm}
\label{tab:binary}
\end{table}

\subsection{Why Does MixSP Work?} \label{subsec:why}

In this experiment, we explain why our method outperforms previous works based on two metrics: (i) the overlap area in sentence embedding and (ii) the upper-range and lower-range alignment scores.

\noindent
\textbf{Reduced Upper-Lower Ranges Overlap}.
%
% Purpose of this paragraph:
% - อธิบายผลการทดลองการแบ่ง 2 คลาสและวิเคราะห์ว่าการลด overlap area สามารถช่วยเพิ่มคะแนน sts เพื่อตอบคำถามที่บอกไว้ใน introsuction
% - รวมไปถึงว่าให้ insight อะไรบ้าง
%
%
%
As discussed in the introduction (Figure~\ref{fig:overlap_area_cd_test}), MixSP produces the smallest upper-lower ranges overlap compared to FT and MoE. 
In this subsection, let us examine how this quantity relates to the overall, upper-range, and lower-range raking performances shown in Figure~\ref{fig:overlap}.
%
%The overlapped areas are displayed in Figure~\ref{fig:overall_overlapped} (X-axis).
%
Results are reported as the average over the three STS datasets used in the main experiment. For the full results, please refer to Table~\ref{tab:all_overlap_area} in the appendix section.
As can be seen, MixSP exhibits the smallest overlap compared to competitive methods.
Figure~\ref{fig:overall_overlapped} shows that we decrease the overlap from 0.360 (SimCSE) and 0.281 (MoE) to 0.265.
Regarding the overall ranking performance, MixSP provides the highest SRC.

\noindent
\textbf{Ranking improvement in upper-range and lower-range classes}.
Figures~\ref{fig:pos} and~\ref{fig:neg} provide an insight into the ranking performance within each class.
We can see that MixSP is also the best performer regarding the upper-range and lower-range ranking.
Interestingly, Figures~\ref{fig:pos} and~\ref{fig:neg} show contrasting results for FT regarding upper-range and lower-range classes ranking performances.
Compared to the SimCSE baseline, FT provides a performance drop for the upper-range class and a performance increase for the lower-range class.
One possible explanation comes from our hypothesis that the upper-range and lower-range samples differ linguistically. 
Since the lower-range class dominates the datasets, familiarizing the model with an STS dataset helps improve the lower-range ranking \emph{but} detriments the upper-range ranking performance.
Another valuable insight obtained from this analysis is the performance gap between upper-range and lower-range classes.
All methods exhibit an upper-lower range performance difference of at least 31.51 points.
This insight suggests that more research attention should be dedicated to improving upper-range raking performance.

\noindent
\textbf{Summary}.
%
% Purpose of this paragraph:
% - สรุปผลการทดลองการแบ่ง 2 คลาสแล้วให้แต่ละคลาสเรียง similarity ของคลาสตัวเองดีกว่ามองเป็นคลาสเดียว
% - เพื่อตอบคำถามที่พูดไว้ใน introduction
%
The space decomposition mechanism in MixSP produces the smallest upper-lower ranges overlap and obtains the best raking performance overall, as well as the individual cases of upper-range and lower-range classes.
These results highlight the benefits of dividing the samples into upper-range and lower-range classes with the assistance of the routing network and specialized projectors. 
Our analysis also shows the ranking performance gap between the upper-range and lower-range classes.
This insight suggests where the research attention should be dedicated to improving the STS ranking performance.

\subsection{Ablation Study} \label{subsec:ablation_study}
%
% Purpose of this paragraph:
% - อธิบายผลการทดลองและวิเคราะห์ว่าทำไมวิธีนี้ถึงทำได้ดีกว่าวิธีอื่นๆ
% - รวมไปถึงว่าให้ insight อะไรบ้าง โดยแบ่งเป็น 3 part ตาม pipeline MixSP คือ routing network, specialized projectors และ trainign objective
%

%
In this study, we analyze our framework's performance and design choice, including the routing network, specialized projectors, and training objectives. 
In addition, we use SimCSE+MixSP as our baseline. 
The analyses of each component are presented as follows. 

\begin{table}[h]
\renewcommand{\arraystretch}{1}
\vspace{-2mm}
\hspace*{-4mm}
\centering
\setlength\doublerulesep{4pt}
\scalebox{0.8}{
\setlength{\tabcolsep}{5pt}
%\fontsize{10pt}{12pt}\selectfont
\begin{tabular}{ll}
\hline
\multicolumn{1}{l|}{\textbf{Method}}                 & \multicolumn{1}{c}{\textbf{BERT-Base}} \\ \hline \hline
\multicolumn{1}{l|}{MixSP}                   & \multicolumn{1}{c}{85.39}   \\ \hline

\multicolumn{2}{l}{\textit{Routing network}}                            \\ \hline
% \multicolumn{1}{l|}{MoE}  & \multicolumn{1}{c}{\color[HTML]{FE0000}$\downarrow$2.15}   \\ 
\multicolumn{1}{l|}{Using only $\mathbf{h}_{\text{x}j}$ for the routing network}  & \multicolumn{1}{c}{\color[HTML]{FE0000}$\downarrow$1.39}   \\ 
\multicolumn{1}{l|}{Using $\mathbf{h}_{\text{x1}}$ and $\mathbf{h}_{\text{x2}}$ for the routing network}  & \multicolumn{1}{c}{\color[HTML]{FE0000}$\downarrow$1.30}   \\ 
\multicolumn{1}{l|}{Removing $\mathcal{L}_{\text{clf}}$} & \multicolumn{1}{c}{\color[HTML]{FE0000}$\downarrow$1.10}                                      \\ \hline
\multicolumn{2}{l}{\textit{Specialized projectors}} \\ \hline
\multicolumn{1}{l|}{[0,4), [4,5]$\rightarrow$[0,1), [1,2), [2,3), [3,4), [4,5]}                     & \multicolumn{1}{c}{\color[HTML]{FE0000}$\downarrow$0.73}    \\ 
\multicolumn{1}{l|}{[0,4), [4,5]$\rightarrow$[0,3), [3,5]}                     & \multicolumn{1}{c}{\color[HTML]{FE0000}$\downarrow$0.92}   \\ 
\multicolumn{1}{l|}{[0,4), [4,5]$\rightarrow$[0,2), [2,5]}                     & \multicolumn{1}{c}{\color[HTML]{FE0000}$\downarrow$1.09}   \\        
\multicolumn{1}{l|}{Argmax$\rightarrow$Weighted-average pooling}                     & 
\multicolumn{1}{c}{\color[HTML]{FE0000}$\downarrow$1.31}   \\ \hline
\multicolumn{2}{l}{\textit{Training objectives}} \\ \hline
\multicolumn{1}{l|}{BCE$\rightarrow$Contrastive learning}                     & \multicolumn{1}{c}{\color[HTML]{FE0000}$\downarrow$3.43}   \\ 
\multicolumn{1}{l|}{BCE$\rightarrow$Cosine similarity}                     & \multicolumn{1}{c}{\color[HTML]{FE0000}$\downarrow$1.79}   \\ \hline \hline
\end{tabular}
}
\vspace{-1mm}
\caption{ The design choice of our framework. We evaluate the average STS score across three STS datasets. $\rightarrow$ is replacing the left method with the right method.}
\vspace{-3mm}
\label{tab:routing_expertise}
\end{table}

\noindent
\textbf{Routing network}. As presented in Table~\ref{tab:routing_expertise}, the best setting of the routing network is the default setting (our decomposition embedding space setting). 
For example, we found performance decreases by 1.10 points when removing the classification loss ($\mathcal{L}_{\text{Clf}}$).
This outcome underscores the importance of having a supervised signal for the router in contrast to the MoE, which lets the attention module automatically decide how to route.
Altering the default setting, which is designed based on the desired property and linguistic observation, harms the model's performance in all cases.  
%
% For example, removing the global contextualize representation $\mathbf{h}_{\text{c}}$ results in the routing network lacking a relation between sentence1 and sentence2, resulting in a 1.05 point performance penalty. 
%

\noindent
\textbf{Specialized projectors}. 
Table~\ref{tab:routing_expertise} shows a consistent decline in the STS score when we change from the default range, [0,4) and [4,5], to other ranges.
Specifically, when changing from the default range to five ranges, there is a noticeable drop in performance by 0.73 points.
In addition, when projector representations are combined through weighted-average pooling similar to MoE, the results decrease by 1.31 points.
%
% This result emphasizes the importance, benefit, and promising direction of splitting the embedding space into two spaces.
%
This trend highlights the effective representation achieved using two projectors separately to sufficiently capture upper-range and lower-range samples.
%
% Separating positive and negative samples is a promising direction.

\noindent
\textbf{Training objective}. One of the key successes of MixSP is the training objective.
In this study, we changed from binary cross-entropy to well-established supervised training objectives, such as contrastive learning and cosine similarity.
As shown in Table~\ref{tab:routing_expertise}, the experimental results demonstrate that using the default training objective yields the best STS score.
While contrastive learning demonstrated a reasonable performance in competitive methods, it adversely impacted the model's performance more than any other setting.
This is because contrastive learning combines negative representations produced from different specialized projectors through a mini-batch negative sample.
Mixing both representations from upper-range and lower-range projectors in the training step only creates confusion between the two classes, deteriorating the model's performance.
The experimental result from Argmax$\rightarrow$Weighted-average pooling conforms with this analysis.

\subsection{Vanilla Vs. End-to-End Routing Networks} \label{subsubsec:routing}
A common technique to separate two classes (i.e., upper-range and lower-range classes) is training a sequence text classification with a PLM~\cite{DBLP:conf/naacl/DevlinCLT19}, while our work trains the routing network simultaneously with the representation learning (the end-to-end manner). 
In this study, we evaluate the effectiveness of our routing network compared to the vanilla text classification model on three STS benchmarks.
Table~\ref{tab:routing_acc} demonstrates that training a text classification in an end-to-end manner outperforms the vanilla model in all metrics.
This is because our routing network's training objective $\mathcal{L}_{\text{Clf}}$ receives the benefit from the $\mathcal{L}_{\text{RL}}$ loss by dynamically adjusting gradients alongside representation learning, thereby enhancing classification performance.
This result confirms that training the group classification in an end-to-end manner surpasses the performance of the two-stage paradigm.

\begin{table}[h!]
\vspace{-1mm}
\hspace*{-4mm}
\centering
\setlength\doublerulesep{4pt}
\scalebox{0.65}{
\setlength{\tabcolsep}{5pt}
%\fontsize{10pt}{12pt}\selectfont
\begin{tabular}{l|c|c|c|c}
\hline
\multicolumn{1}{c|}{\textbf{Model}} &
  \textbf{BIOSSES} &
  \textbf{CDSC$\textbf{-}$R(Val)} &
  \textbf{CDSC$\textbf{-}$R(Test)} &
  \textbf{Avg.} \\ \hline \hline
End\textendash to\textendash end       & \textbf{95.50}       & \textbf{88.80}     & \textbf{86.95}     & \textbf{90.42}     \\ 
Vanilla     & 94.00     & 82.60     & 82.40     & 86.33  \\ \hline \hline
\end{tabular}
}
\vspace{-2mm}
\caption{The accuracy score of our routing network (end-to-end) compared to a vanilla text classification model on three STS benchmarks.}
\vspace{-2mm}
\label{tab:routing_acc}
\end{table}

\subsection{Tuning Time and Memory Requirements} \label{subsubsec:inference_speed}
%
% Purpose of this paragraph:
% - เนื่องจากวิธีของเราทำให้ขนาดโมเดลเพิ่มขึ้นจาก 110M เป็น 145M
% - ดังนั้น ย่อหน้านี้ต้องการที่จะอธิบายว่าถึงแม้ขนาดโมเดลเพิ่มขึ้นแต่การใช้ resource และ training time ไม่ได้เพิ่มตาม
%

%
% One of the main concerns in this work is how much of MixSE's parameters and speed efficiency compared to competitive methods.
% %
% we evaluate our method's parameter changes and training time usage to answer the concern.
% %
Table~\ref{tab:time} compares the original SimCSE and three competitive tuning methods in terms of the number of parameters, tuning time, GPU memory consumption (during tuning), and the SRC score.
Since FT \emph{does not} introduce any new component, it has the shortest tuning time and consumes the smallest GPU memory. 
MoE and MixSP incur substantially higher costs than FT, which are comparable to each other, while MixSP provides the highest performance uplift among the three tuning methods. 

%although the parameters number of MixSP increased from 110 to 145 million, we improved Spearman's rank correlation (SRC) from 69.74 to 85.39 points.
%
%Moreover, we found that the training time and GPU memory usage of MixSP are comparable to MoE, while the performance gap between MixSP and MoE is 4.30 points.
%

\begin{table}[h!]
\renewcommand{\arraystretch}{1}
\vspace{-2mm}

\hspace*{-3mm}
\centering
\setlength\doublerulesep{5pt}
\scalebox{0.80}{
\setlength{\tabcolsep}{2pt}
%\fontsize{10pt}{12pt}\selectfont
\begin{tabular}{l|c|c|c|c}
\hline
\textbf{Method} & \textbf{\#Params} & \textbf{Tuning time}  & \textbf{GPU memory} & \textbf{SRC Avg.}\\ \hline \hline
SimCSE          & \textbf{110M}     & -                & -   & 69.73                \\ 
+FT         & \textbf{110M}              & \textbf{83 sec.}                & \textbf{5.606 MBs}      & 72.04           \\ 
+MoE         & 133M              & 182 sec.                & 8,352 MBs      & 81.09           \\ 
+MixSP            & 145M              & 198 sec.  & 8,390 MBs   & \textbf{85.39} \\ \hline \hline
\end{tabular}
}
\vspace{-2mm}
\caption{The number of parameters, training time, GPU memory usage, and the average Spearman's rank correlation (SRC) score from Table~\ref{tab:sts}. We use the same training data, learning rate, epoch, and batch size for the competitive methods and ours.}
\vspace{-1mm}
\label{tab:time}
\end{table}

%% file: 6_Conclusion.tex
\vspace{-2mm}
\section{Conclusion}
\vspace{-2mm}

In this paper, we proposed a novel embedding space decomposition method called MixSP, a mixture of specialized projectors.
We challenge the common practice in treating STS scores from a continuous paradigm [0,5] to embedding space decomposition for lower-range [0,4) and upper-range [4,5] classes.   
Our experiments highlight that MixSP outperforms competitive methods in the average cases of STS and zero-shot benchmarks.
We also discuss the improvement of our method, design choice, and inference speed to emphasize the effectiveness of our framework.
The embedding space decomposition is the promising paradigm for sentence embedding. 
%
% emphasizing the significance of design choices, including divided embedding spaces and varied training objectives.
%
% Our method provides a new perspective on the STS problem by segregating positive and negative classes into embedding decomposition, integrated into language model pre-training.

\section{Limitation}
Since our method incorporates a learnable module for enhancing sentence embeddings, MixSP's parameter increases from 110 million to 145 million parameters, as shown in Table~\ref{tab:time}. 
In addition, the training time and memory usage of our method are also higher than competitor methods.
However, regarding performance, our method outperforms competitors in all cases.

%% file: 7_Appendix.tex
\section{Appendix}

% \subsection{Overlap In Detail} \label{appendix:overlap}
% %
% Let us now consider cosine similarity distributions for the STS score ranges of  [0,4) (the rank score of irrelevance sentence pairs) and  [4,5] (the rank score of relevance sentence pairs) using the STS-B development dataset.
% %
% Figure~\ref{fig:overlap_area_sts15} (in the introduction) shows distribution plots for BERT-Base produced from our method and competitive methods.
% %
% We derive a confusion-matrix-like measure to provide more insight into the behavior of our embedding space.
% %
% In particular, we compute the overlapped density area by estimating the densities using the Gaussian kernel density estimate.
% %
% This measure indicates the chance that sentence pairs from the two ranges can be confused with each other.
% %
% We want the overlapped areas to be as close to zero as possible.
% %
% For the full overlap score, please refer to Table~\ref{tab:all_overlap_area}.

\label{sec:appendix}

\begin{table*}[h!]
\vspace{-1mm}
\hspace*{-4mm}
\centering
\setlength\doublerulesep{4pt}
\scalebox{0.8}{
\setlength{\tabcolsep}{5pt}
%\fontsize{10pt}{12pt}\selectfont
\begin{tabular}{l|c|c|c}
\hline
\multicolumn{1}{c|}{\textbf{Dataset}} &
  \textbf{4$\textbf{-}$gram similarity} &
  \textbf{5$\textbf{-}$gram similarity} &
  \textbf{6$\textbf{-}$gram similarity} \\ \hline \hline
STS12   & 0.1168     & 0.1059     & 0.0968     \\ 
STS13   & 0.0147     & 0.0086     & 0.0048     \\ 
STS14   & 0.1492     & 0.1416     & 0.1876     \\
STS15   & 0.0675     & 0.0611     & 0.0546     \\
STS16   & 0.0066     & 0.0040     & 0.0024     \\
SICK\textendash R  & 0.0242     & 0.0160     & 0.0100     \\
STS\textendash B  & 0.0258     & 0.0188     & 0.0148     \\
BIOSSES   & 0.0001     & \textendash     & \textendash             \\
CDSC\textendash R (validation set)   & \textendash     & \textendash     & \textendash             \\
CDSC\textendash R (test set)   & \textendash     & \textendash     & \textendash             \\
\hline \hline
\end{tabular}
}
\vspace{-2mm}
\caption{\label{font-table} The comparison of N-gram similarity between the STS-B training set and other STS corpora.}
\vspace{-2mm}
\label{tab:n_gram_sim}
\end{table*}

\subsection{Competitive Method Implementations} \label{appendix:competitive_methods}
As stated in Section~\ref{subsec:comp}, we compare our method with two competitive methods, such as +FT and +MoE.
We explain in more detail about the implementation and differences between our work and competitive methods as follows:
\begin{compactitem}[\hspace{\setalign}•]
    \item \textbf{+FT} [No Space Decomposition]: First, we modify the model architecture to conform with the learning process described in Sentence Transformers~\cite{reimers-2019-sentence-bert}. Then, for fine-tuning, we utilize the cosine similarity loss function to train the model, comparing predicted and ground truth similarity scores. The final result is a model that takes pairs of sentences and outputs similarity scores for STS. FT treats the entire STS score range [0,5], providing a contrast to our method, which separates the ranges [0,4) and [4,5].

    \item \textbf{+MoE} [Soft Selection]: First, we utilize the standard cross-encoder architecture~\cite{reimers-2019-sentence-bert}. Second, we modify the model to incorporate the MoE technique~\cite{NEURIPS2022_2f00ecd7} and train the model using the BCE loss function to compare predicted and ground truth similarity scores. Similar to FT, we obtain a model that accepts a sentence pair and outputs a similarity score as the final result. While MoE decomposes the embedding space using multiple experts, they combine outputs using weighted average or soft-selection. In contrast, MixSP performs a hard selection on the output, utilizing only one output representation at a time.
    %
    % Additionally, the routing network includes a supervised signal to enhance the model's understanding of sentence relationships.

\end{compactitem}

\subsection{Data Leakage} \label{appendix:data_leakage}
In this study, we demonstrate the data leakage in STS-B and standard STS benchmarks.
We found that STS-B has a high n-gram overlap with the test data (STS 12-16, SICK-R, and STS-B test set). 
This is because STS-B comprises samples from STS datasets between 2012 to 2016, as stated on the STS-B's website~\footnote{\url{https://ixa2.si.ehu.eus/stswiki/stswiki/index.php/Special:Random.html}}. 
In addition, we also conducted an n-gram overlap analysis and found a high Jaccard similarity compared to other datasets, as shown in Table~\ref{tab:n_gram_sim}.
Therefore, we need to omit the high word-overlap dataset from our main experimental results.
Our experimental results only consisted of the CDSR-R and BIOSSES datasets.

\subsection{Seven Standard STS Datasets} \label{subsec:sts_standard}
In this study, we evaluate the effectiveness of our method and competitive methods on the traditional seven benchmarks.
Note that all models were trained on STS-B training data.
However, we notice that using the STS-B as the training data might cause data leakage, as discussed in Appendix~\ref{appendix:data_leakage}. 
Thus, we did not include these results in the main paper.

As shown in Table~\ref{tab:base_encoder}, our method outperforms competitive methods on the average score.
We observe performance improvements compared to base encodes in all cases for all methods.
Moreover, we also notice the performance gap in SBERT on the main table (Table~\ref{tab:sts}) and seven standard STS datasets.
This is because the data leakage setting and SBERT seem to find a shortcut in this setting. 

\begin{table*}[h]
\vspace{-1mm}
\hspace*{-4mm}
\centering
\setlength\doublerulesep{3pt}
\scalebox{0.8}{
\setlength{\tabcolsep}{8pt}
%\fontsize{10pt}{12pt}\selectfont
\begin{tabular}{l|cccccccc}
\hline
\multicolumn{1}{l|}{\textbf{Method}} &
  \multicolumn{1}{l|}{\textbf{STS12}} &
  \multicolumn{1}{l|}{\textbf{STS13}} &
  \multicolumn{1}{l|}{\textbf{STS14}} &
  \multicolumn{1}{l|}{\textbf{STS15}} &
  \multicolumn{1}{l|}{\textbf{STS16}} &
  \multicolumn{1}{l|}{\textbf{STS-B}} &
  \multicolumn{1}{l|}{\textbf{SICK-R}} &
  \multicolumn{1}{l}{\textbf{Avg.}} \\ \hline \hline
\multicolumn{9}{c}{\textit{SBERT as the base encoder}} \\ \hline
SBERT &
  \multicolumn{1}{c|}{72.51} &
  \multicolumn{1}{c|}{87.16} &
  \multicolumn{1}{c|}{84.32} &
  \multicolumn{1}{c|}{85.26} &
  \multicolumn{1}{c|}{79.65} &
  \multicolumn{1}{c|}{82.67} &
  \multicolumn{1}{c|}{73.37} &
  80.71 \\
+MoE &
  \multicolumn{1}{c|}{\textbf{80.71}} &
  \multicolumn{1}{c|}{87.89} &
  \multicolumn{1}{c|}{89.88} &
  \multicolumn{1}{c|}{90.83} &
  \multicolumn{1}{c|}{83.75} &
  \multicolumn{1}{c|}{86.14} &
  \multicolumn{1}{c|}{75.01} &
  84.89 \\
+MixSP &
  \multicolumn{1}{c|}{79.99} &
  \multicolumn{1}{c|}{\textbf{89.47}} &
  \multicolumn{1}{c|}{\textbf{89.99}} &
  \multicolumn{1}{c|}{\textbf{90.89}} &
  \multicolumn{1}{c|}{\textbf{84.59}} &
  \multicolumn{1}{c|}{\textbf{87.27}} &
  \multicolumn{1}{c|}{\textbf{76.23}} &
  \textbf{85.49} \\ \hline
\multicolumn{9}{c}{\textit{SimCSE as the base encoder}} \\ \hline
SimCSE &
  \multicolumn{1}{c|}{75.30} &
  \multicolumn{1}{c|}{84.67} &
  \multicolumn{1}{c|}{80.19} &
  \multicolumn{1}{c|}{85.40} &
  \multicolumn{1}{c|}{80.82} &
  \multicolumn{1}{c|}{84.25} &
  \multicolumn{1}{c|}{80.39} &
  81.57 \\
+FT &
  \multicolumn{1}{c|}{79.15} &
  \multicolumn{1}{c|}{89.44} &
  \multicolumn{1}{c|}{89.33} &
  \multicolumn{1}{c|}{90.46} &
  \multicolumn{1}{c|}{83.50} &
  \multicolumn{1}{c|}{87.04} &
  \multicolumn{1}{c|}{\textbf{80.40}} &
  85.39 \\
+MoE &
  \multicolumn{1}{c|}{79.38} &
  \multicolumn{1}{c|}{89.02} &
  \multicolumn{1}{c|}{89.80} &
  \multicolumn{1}{c|}{\textbf{91.35}} &
  \multicolumn{1}{c|}{83.38} &
  \multicolumn{1}{c|}{86.34} &
  \multicolumn{1}{c|}{76.43} &
  85.10 \\
+MixSP &
  \multicolumn{1}{c|}{\textbf{81.08}} &
  \multicolumn{1}{c|}{\textbf{89.74}} &
  \multicolumn{1}{c|}{\textbf{90.41}} &
  \multicolumn{1}{c|}{91.27} &
  \multicolumn{1}{c|}{\textbf{84.16}} &
  \multicolumn{1}{c|}{\textbf{87.18}} &
  \multicolumn{1}{c|}{75.86} &
  \textbf{85.67} \\ \hline
\multicolumn{9}{c}{\textit{DiffAug as the base encoder}} \\ \hline
DiffAug &
  \multicolumn{1}{c|}{76.92} &
  \multicolumn{1}{c|}{85.17} &
  \multicolumn{1}{c|}{80.81} &
  \multicolumn{1}{c|}{86.91} &
  \multicolumn{1}{c|}{82.52} &
  \multicolumn{1}{c|}{84.32} &
  \multicolumn{1}{c|}{\textbf{80.27}} &
  82.42 \\
+FT &
  \multicolumn{1}{c|}{77.30} &
  \multicolumn{1}{c|}{89.56} &
  \multicolumn{1}{c|}{87.77} &
  \multicolumn{1}{c|}{88.54} &
  \multicolumn{1}{c|}{82.02} &
  \multicolumn{1}{c|}{85.36} &
  \multicolumn{1}{c|}{78.94} &
  84.21 \\
+MoE &
  \multicolumn{1}{c|}{80.18} &
  \multicolumn{1}{c|}{88.74} &
  \multicolumn{1}{c|}{\textbf{90.25}} &
  \multicolumn{1}{c|}{\textbf{91.48}} &
  \multicolumn{1}{c|}{83.63} &
  \multicolumn{1}{c|}{86.29} &
  \multicolumn{1}{c|}{76.19} &
  85.25 \\
+MixSP &
  \multicolumn{1}{c|}{\textbf{80.88}} &
  \multicolumn{1}{c|}{\textbf{88.89}} &
  \multicolumn{1}{c|}{89.64} &
  \multicolumn{1}{c|}{91.46} &
  \multicolumn{1}{c|}{\textbf{84.53}} &
  \multicolumn{1}{c|}{\textbf{87.00}} &
  \multicolumn{1}{c|}{75.88} &
  \textbf{85.47} \\ \hline\hline
\end{tabular}
}
\vspace{-2mm}
\caption{Spearman’s rank correlation on the seven STS benchmarks~\cite{cer-etal-2017-semeval, marelli-etal-2014-sick, agirre-etal-2012-semeval,agirre-etal-2013-sem,agirre-etal-2014-semeval,agirre-etal-2015-semeval,agirre-etal-2016-semeval}. All models were implemented on BERT-Base.}
\vspace{-2mm}
\label{tab:base_encoder}
\end{table*}

\begin{table*}[ht!]
\vspace{-1mm}
\hspace*{-4mm}
\centering
\setlength\doublerulesep{4pt}
\scalebox{0.8}{
\setlength{\tabcolsep}{5pt}
%\fontsize{10pt}{12pt}\selectfont
\begin{tabular}{l|c|c|c|c|c|c|c|c|c|c|c|c}
\hline
\multicolumn{1}{c|}{\multirow{2}{*}{\textbf{Model}}} &
  \multicolumn{3}{c|}{\textbf{BIOSSES}} &
  \multicolumn{3}{c|}{\textbf{CDSC$\textbf{-}$R(Val)}} &
  \multicolumn{3}{c|}{\textbf{CDSC$\textbf{-}$R(Test)}} &
  \multicolumn{3}{c}{\textbf{Avg.}} \\ \cline{2-13} 
 &
  \textbf{Upper} &
  \textbf{Lower} &
  \textbf{Overlap} & 
  \textbf{Upper} &
  \textbf{Lower} &
  \textbf{Overlap} &  
  \textbf{Upper} &
  \textbf{Lower} &
  \textbf{Overlap} & 
  \textbf{Upper} &
  \textbf{Lower} &
  \textbf{Overlap} \\ \hline \hline
SimCSE &
  66.34 & 27.51 & 0.414 &
  56.92 & 30.47 & 0.292 &
  53.78 & 24.53 & 0.374 &
  59.01 & 27.50 & 0.360 \\ 
$+$FT &
  74.12 & 25.22 & 0.234 &
  55.79 & 29.89 & 0.295 &
  53.75 & 20.52 & 0.385 &
  61.28 & 25.21 & 0.305 \\ 
$+$MoE &
  74.21 & 34.36 & 0.238 &
  75.97 & 45.08 & 0.281 &
  79.42 & 23.65 & 0.324 &
  76.53 & 34.37 & 0.281 \\ 
$+$MixSP &
  \textbf{80.06} & \textbf{36.27} & \textbf{0.201} &
  \textbf{79.57} & 44.69 & \textbf{0.280} &
  \textbf{83.44} & \textbf{27.83} & \textbf{0.314} &
  \textbf{81.02} & \textbf{36.26} & \textbf{0.265} \\ 
\hline \hline
\end{tabular}
}
\vspace{-2mm}
\caption{\label{font-table} Spearman's rank correlation on the STS benchmarks where Upper is upper-range samples: STS $\in [4,5]$, Lower is lower-range samples: STS $\in [0,4)$, and Overlap is the overlap area between $[0,4)$ and $[4,5]$.}
\vspace{-2mm}
\label{tab:all_overlap_area}
\end{table*}

%% file: main.bbl
\begin{thebibliography}{37}
\expandafter\ifx\csname natexlab\endcsname\relax\def\natexlab#1{#1}\fi

\bibitem[{Agirre et~al.(2015)Agirre, Banea, Cardie, Cer, Diab, Gonzalez-Agirre, Guo, Lopez-Gazpio, Maritxalar, Mihalcea, Rigau, Uria, and Wiebe}]{agirre-etal-2015-semeval}
Eneko Agirre, Carmen Banea, Claire Cardie, Daniel Cer, Mona Diab, Aitor Gonzalez-Agirre, Weiwei Guo, I{\~n}igo Lopez-Gazpio, Montse Maritxalar, Rada Mihalcea, German Rigau, Larraitz Uria, and Janyce Wiebe. 2015.
\newblock \href {https://doi.org/10.18653/v1/S15-2045} {{S}em{E}val-2015 task 2: Semantic textual similarity, {E}nglish, {S}panish and pilot on interpretability}.
\newblock In \emph{Proceedings of the 9th International Workshop on Semantic Evaluation ({S}em{E}val 2015)}, pages 252--263, Denver, Colorado. Association for Computational Linguistics.

\bibitem[{Agirre et~al.(2014)Agirre, Banea, Cardie, Cer, Diab, Gonzalez-Agirre, Guo, Mihalcea, Rigau, and Wiebe}]{agirre-etal-2014-semeval}
Eneko Agirre, Carmen Banea, Claire Cardie, Daniel Cer, Mona Diab, Aitor Gonzalez-Agirre, Weiwei Guo, Rada Mihalcea, German Rigau, and Janyce Wiebe. 2014.
\newblock \href {https://doi.org/10.3115/v1/S14-2010} {{S}em{E}val-2014 task 10: Multilingual semantic textual similarity}.
\newblock In \emph{Proceedings of the 8th International Workshop on Semantic Evaluation ({S}em{E}val 2014)}, pages 81--91, Dublin, Ireland. Association for Computational Linguistics.

\bibitem[{Agirre et~al.(2016)Agirre, Banea, Cer, Diab, Gonzalez-Agirre, Mihalcea, Rigau, and Wiebe}]{agirre-etal-2016-semeval}
Eneko Agirre, Carmen Banea, Daniel Cer, Mona Diab, Aitor Gonzalez-Agirre, Rada Mihalcea, German Rigau, and Janyce Wiebe. 2016.
\newblock \href {https://doi.org/10.18653/v1/S16-1081} {{S}em{E}val-2016 task 1: Semantic textual similarity, monolingual and cross-lingual evaluation}.
\newblock In \emph{Proceedings of the 10th International Workshop on Semantic Evaluation ({S}em{E}val-2016)}, pages 497--511, San Diego, California. Association for Computational Linguistics.

\bibitem[{Agirre et~al.(2012)Agirre, Cer, Diab, and Gonzalez-Agirre}]{agirre-etal-2012-semeval}
Eneko Agirre, Daniel Cer, Mona Diab, and Aitor Gonzalez-Agirre. 2012.
\newblock \href {https://aclanthology.org/S12-1051} {{S}em{E}val-2012 task 6: A pilot on semantic textual similarity}.
\newblock In \emph{*{SEM} 2012: The First Joint Conference on Lexical and Computational Semantics {--} Volume 1: Proceedings of the main conference and the shared task, and Volume 2: Proceedings of the Sixth International Workshop on Semantic Evaluation ({S}em{E}val 2012)}, pages 385--393, Montr{\'e}al, Canada. Association for Computational Linguistics.

\bibitem[{Agirre et~al.(2013)Agirre, Cer, Diab, Gonzalez-Agirre, and Guo}]{agirre-etal-2013-sem}
Eneko Agirre, Daniel Cer, Mona Diab, Aitor Gonzalez-Agirre, and Weiwei Guo. 2013.
\newblock \href {https://aclanthology.org/S13-1004} {*{SEM} 2013 shared task: Semantic textual similarity}.
\newblock In \emph{Second Joint Conference on Lexical and Computational Semantics (*{SEM}), Volume 1: Proceedings of the Main Conference and the Shared Task: Semantic Textual Similarity}, pages 32--43, Atlanta, Georgia, USA. Association for Computational Linguistics.

\bibitem[{Bowman et~al.(2015)Bowman, Angeli, Potts, and Manning}]{bowman-etal-2015-large}
Samuel~R. Bowman, Gabor Angeli, Christopher Potts, and Christopher~D. Manning. 2015.
\newblock \href {https://doi.org/10.18653/v1/D15-1075} {A large annotated corpus for learning natural language inference}.
\newblock In \emph{Proceedings of the 2015 Conference on Empirical Methods in Natural Language Processing}, pages 632--642, Lisbon, Portugal. Association for Computational Linguistics.

\bibitem[{Cer et~al.(2017)Cer, Diab, Agirre, Lopez-Gazpio, and Specia}]{cer-etal-2017-semeval}
Daniel Cer, Mona Diab, Eneko Agirre, I{\~n}igo Lopez-Gazpio, and Lucia Specia. 2017.
\newblock \href {https://doi.org/10.18653/v1/S17-2001} {{S}em{E}val-2017 task 1: Semantic textual similarity multilingual and crosslingual focused evaluation}.
\newblock In \emph{Proceedings of the 11th International Workshop on Semantic Evaluation ({S}em{E}val-2017)}, pages 1--14, Vancouver, Canada. Association for Computational Linguistics.

\bibitem[{Chen et~al.(2023)Chen, Shen, Ding, Chen, Zhao, Learned{-}Miller, and Gan}]{DBLP:conf/cvpr/ChenSDCZLG23}
Zitian Chen, Yikang Shen, Mingyu Ding, Zhenfang Chen, Hengshuang Zhao, Erik~G. Learned{-}Miller, and Chuang Gan. 2023.
\newblock \href {https://doi.org/10.1109/CVPR52729.2023.01138} {Mod-squad: Designing mixtures of experts as modular multi-task learners}.
\newblock In \emph{{IEEE/CVF} Conference on Computer Vision and Pattern Recognition, {CVPR} 2023, Vancouver, BC, Canada, June 17-24, 2023}, pages 11828--11837. {IEEE}.

\bibitem[{Chowdhury et~al.(2023)Chowdhury, Zhang, Wang, Liu, and Chen}]{DBLP:conf/icml/ChowdhuryZW0C23}
Mohammed Nowaz~Rabbani Chowdhury, Shuai Zhang, Meng Wang, Sijia Liu, and Pin{-}Yu Chen. 2023.
\newblock \href {https://proceedings.mlr.press/v202/chowdhury23a.html} {Patch-level routing in mixture-of-experts is provably sample-efficient for convolutional neural networks}.
\newblock In \emph{International Conference on Machine Learning, {ICML} 2023, 23-29 July 2023, Honolulu, Hawaii, {USA}}, volume 202 of \emph{Proceedings of Machine Learning Research}, pages 6074--6114. {PMLR}.

\bibitem[{Chuang et~al.(2022)Chuang, Dangovski, Luo, Zhang, Chang, Soljacic, Li, Yih, Kim, and Glass}]{chuang2022diffcse}
Yung-Sung Chuang, Rumen Dangovski, Hongyin Luo, Yang Zhang, Shiyu Chang, Marin Soljacic, Shang-Wen Li, Wen-tau Yih, Yoon Kim, and James Glass. 2022.
\newblock {DiffCSE}: Difference-based contrastive learning for sentence embeddings.
\newblock In \emph{Annual Conference of the North American Chapter of the Association for Computational Linguistics (NAACL)}.

\bibitem[{Devlin et~al.(2019)Devlin, Chang, Lee, and Toutanova}]{DBLP:conf/naacl/DevlinCLT19}
Jacob Devlin, Ming{-}Wei Chang, Kenton Lee, and Kristina Toutanova. 2019.
\newblock \href {https://doi.org/10.18653/v1/n19-1423} {{BERT:} pre-training of deep bidirectional transformers for language understanding}.
\newblock In \emph{{NAACL-HLT} 2019}.

\bibitem[{Fang et~al.(2020)Fang, Wang, Zhou, Ding, and Xie}]{fang2020cert}
Hongchao Fang, Sicheng Wang, Meng Zhou, Jiayuan Ding, and Pengtao Xie. 2020.
\newblock \href {http://arxiv.org/abs/2005.12766} {Cert: Contrastive self-supervised learning for language understanding}.

\bibitem[{Gao et~al.(2021)Gao, Yao, and Chen}]{DBLP:conf/emnlp/GaoYC21}
Tianyu Gao, Xingcheng Yao, and Danqi Chen. 2021.
\newblock \href {https://aclanthology.org/2021.emnlp-main.552} {Simcse: Simple contrastive learning of sentence embeddings}.
\newblock In \emph{{EMNLP} 2021}.

\bibitem[{Jiang et~al.(2022{\natexlab{a}})Jiang, Jiao, Huang, Zhang, Wang, Zhuang, Wei, Huang, Deng, and Zhang}]{DBLP:conf/emnlp/JiangJHZWZWHDZ22}
Ting Jiang, Jian Jiao, Shaohan Huang, Zihan Zhang, Deqing Wang, Fuzhen Zhuang, Furu Wei, Haizhen Huang, Denvy Deng, and Qi~Zhang. 2022{\natexlab{a}}.
\newblock \href {https://doi.org/10.18653/v1/2022.emnlp-main.603} {Promptbert: Improving {BERT} sentence embeddings with prompts}.
\newblock In \emph{Proceedings of the 2022 Conference on Empirical Methods in Natural Language Processing, {EMNLP} 2022, Abu Dhabi, United Arab Emirates, December 7-11, 2022}, pages 8826--8837. Association for Computational Linguistics.

\bibitem[{Jiang et~al.(2022{\natexlab{b}})Jiang, Zhang, and Wang}]{jiang-etal-2022-improved}
Yuxin Jiang, Linhan Zhang, and Wei Wang. 2022{\natexlab{b}}.
\newblock \href {https://aclanthology.org/2022.findings-emnlp.220} {Improved universal sentence embeddings with prompt-based contrastive learning and energy-based learning}.
\newblock In \emph{Findings of the Association for Computational Linguistics: EMNLP 2022}, pages 3021--3035, Abu Dhabi, United Arab Emirates. Association for Computational Linguistics.

\bibitem[{Li et~al.(2023)Li, Shen, Yang, Wang, Ren, Che, Zhang, and Liu}]{DBLP:conf/iclr/LiSYWRCZ023}
Bo~Li, Yifei Shen, Jingkang Yang, Yezhen Wang, Jiawei Ren, Tong Che, Jun Zhang, and Ziwei Liu. 2023.
\newblock \href {https://openreview.net/pdf?id=RecZ9nB9Q4} {Sparse mixture-of-experts are domain generalizable learners}.
\newblock In \emph{The Eleventh International Conference on Learning Representations, {ICLR} 2023, Kigali, Rwanda, May 1-5, 2023}. OpenReview.net.

\bibitem[{Li et~al.(2020)Li, Zhou, He, Wang, Yang, and Li}]{DBLP:conf/emnlp/LiZHWYL20}
Bohan Li, Hao Zhou, Junxian He, Mingxuan Wang, Yiming Yang, and Lei Li. 2020.
\newblock \href {https://doi.org/10.18653/V1/2020.EMNLP-MAIN.733} {On the sentence embeddings from pre-trained language models}.
\newblock In \emph{Proceedings of the 2020 Conference on Empirical Methods in Natural Language Processing, {EMNLP} 2020, Online, November 16-20, 2020}, pages 9119--9130. Association for Computational Linguistics.

\bibitem[{Limkonchotiwat et~al.(2022)Limkonchotiwat, Ponwitayarat, Lowphansirikul, Udomcharoenchaikit, Chuangsuwanich, and Nutanong}]{limkonchotiwat-etal-2022-congen}
Peerat Limkonchotiwat, Wuttikorn Ponwitayarat, Lalita Lowphansirikul, Can Udomcharoenchaikit, Ekapol Chuangsuwanich, and Sarana Nutanong. 2022.
\newblock \href {https://aclanthology.org/2022.findings-emnlp.483} {{C}on{G}en: Unsupervised control and generalization distillation for sentence representation}.
\newblock In \emph{Findings of the Association for Computational Linguistics: EMNLP 2022}, pages 6467--6480, Abu Dhabi, United Arab Emirates. Association for Computational Linguistics.

\bibitem[{Limkonchotiwat et~al.(2023)Limkonchotiwat, Ponwitayarat, Lowphansirikul, Udomcharoenchaikit, Chuangsuwanich, and Nutanong}]{limkonchotiwat-etal-2023-sct}
Peerat Limkonchotiwat, Wuttikorn Ponwitayarat, Lalita Lowphansirikul, Can Udomcharoenchaikit, Ekapol Chuangsuwanich, and Sarana Nutanong. 2023.
\newblock An efficient self-supervised cross-view training for sentence embedding.
\newblock \emph{Transactions of the Association for Computational Linguistics}.

\bibitem[{Liu et~al.(2022)Liu, Jiao, Massiah, Yilmaz, and Havrylov}]{DBLP:conf/iclr/0001JMYH22}
Fangyu Liu, Yunlong Jiao, Jordan Massiah, Emine Yilmaz, and Serhii Havrylov. 2022.
\newblock \href {https://openreview.net/forum?id=AmUhwTOHgm} {Trans-encoder: Unsupervised sentence-pair modelling through self- and mutual-distillations}.
\newblock In \emph{The Tenth International Conference on Learning Representations, {ICLR} 2022, Virtual Event, April 25-29, 2022}. OpenReview.net.

\bibitem[{Liu et~al.(2019)Liu, Ott, Goyal, Du, Joshi, Chen, Levy, Lewis, Zettlemoyer, and Stoyanov}]{DBLP:journals/corr/abs-1907-11692}
Yinhan Liu, Myle Ott, Naman Goyal, Jingfei Du, Mandar Joshi, Danqi Chen, Omer Levy, Mike Lewis, Luke Zettlemoyer, and Veselin Stoyanov. 2019.
\newblock \href {http://arxiv.org/abs/1907.11692} {Roberta: {A} robustly optimized {BERT} pretraining approach}.
\newblock \emph{CoRR}.

\bibitem[{Marelli et~al.(2014)Marelli, Menini, Baroni, Bentivogli, Bernardi, and Zamparelli}]{marelli-etal-2014-sick}
Marco Marelli, Stefano Menini, Marco Baroni, Luisa Bentivogli, Raffaella Bernardi, and Roberto Zamparelli. 2014.
\newblock \href {http://www.lrec-conf.org/proceedings/lrec2014/pdf/363_Paper.pdf} {A {SICK} cure for the evaluation of compositional distributional semantic models}.
\newblock In \emph{Proceedings of the Ninth International Conference on Language Resources and Evaluation ({LREC}'14)}, pages 216--223, Reykjavik, Iceland. European Language Resources Association (ELRA).

\bibitem[{Miao et~al.(2024)Miao, Wu, Zhao, Wu, and Tsuruoka}]{miao2024enhancing}
Zhongtao Miao, Qiyu Wu, Kaiyan Zhao, Zilong Wu, and Yoshimasa Tsuruoka. 2024.
\newblock \href {http://arxiv.org/abs/2404.02490} {Enhancing cross-lingual sentence embedding for low-resource languages with word alignment}.

\bibitem[{Muennighoff et~al.(2023)Muennighoff, Tazi, Magne, and Reimers}]{muennighoff-etal-2023-mteb}
Niklas Muennighoff, Nouamane Tazi, Loic Magne, and Nils Reimers. 2023.
\newblock \href {https://aclanthology.org/2023.eacl-main.148} {{MTEB}: Massive text embedding benchmark}.
\newblock In \emph{Proceedings of the 17th Conference of the European Chapter of the Association for Computational Linguistics}, pages 2014--2037, Dubrovnik, Croatia. Association for Computational Linguistics.

\bibitem[{Opitz and Frank(2022)}]{DBLP:conf/ijcnlp/OpitzF22}
Juri Opitz and Anette Frank. 2022.
\newblock \href {https://aclanthology.org/2022.aacl-main.48} {{SBERT} studies meaning representations: Decomposing sentence embeddings into explainable semantic features}.
\newblock In \emph{Proceedings of the 2nd Conference of the Asia-Pacific Chapter of the Association for Computational Linguistics and the 12th International Joint Conference on Natural Language Processing, {AACL/IJCNLP} 2022 - Volume 1: Long Papers, Online Only, November 20-23, 2022}, pages 625--638. Association for Computational Linguistics.

\bibitem[{Reimers and Gurevych(2019)}]{reimers-2019-sentence-bert}
Nils Reimers and Iryna Gurevych. 2019.
\newblock \href {https://arxiv.org/abs/1908.10084} {Sentence-bert: Sentence embeddings using siamese bert-networks}.
\newblock In \emph{Proceedings of the 2019 Conference on Empirical Methods in Natural Language Processing}. Association for Computational Linguistics.

\bibitem[{Soğancıoğlu et~al.(2017)Soğancıoğlu, Öztürk, and Özgür}]{10.1093/bioinformatics/btx238}
Gizem Soğancıoğlu, Hakime Öztürk, and Arzucan Özgür. 2017.
\newblock \href {https://doi.org/10.1093/bioinformatics/btx238} {{BIOSSES: a semantic sentence similarity estimation system for the biomedical domain}}.
\newblock \emph{Bioinformatics}, 33(14):i49--i58.

\bibitem[{Wang et~al.(2020)Wang, Chen, Wang, and Kuo}]{DBLP:conf/icpr/WangCWK20}
Bin Wang, Fenxiao Chen, Yuncheng Wang, and C.{-}C.~Jay Kuo. 2020.
\newblock \href {https://doi.org/10.1109/ICPR48806.2021.9412169} {Efficient sentence embedding via semantic subspace analysis}.
\newblock In \emph{25th International Conference on Pattern Recognition, {ICPR} 2020, Virtual Event / Milan, Italy, January 10-15, 2021}, pages 119--125. {IEEE}.

\bibitem[{Wang et~al.(2021)Wang, Reimers, and Gurevych}]{wang-etal-2021-tsdae-using}
Kexin Wang, Nils Reimers, and Iryna Gurevych. 2021.
\newblock \href {https://doi.org/10.18653/v1/2021.findings-emnlp.59} {{TSDAE}: Using transformer-based sequential denoising auto-encoderfor unsupervised sentence embedding learning}.
\newblock In \emph{Findings of the Association for Computational Linguistics: EMNLP 2021}, pages 671--688, Punta Cana, Dominican Republic. Association for Computational Linguistics.

\bibitem[{Wang and Lu(2022)}]{wang2022diffaug}
Tianduo Wang and Wei Lu. 2022.
\newblock Differentiable data augmentation for contrastive sentence representation learning.
\newblock In \emph{Empirical Methods in Natural Language Processing (EMNLP)}.

\bibitem[{Wang et~al.(2022)Wang, Xu, Sun, Hu, Tao, Geng, and Jiang}]{wang-etal-2022-promda}
Yufei Wang, Can Xu, Qingfeng Sun, Huang Hu, Chongyang Tao, Xiubo Geng, and Daxin Jiang. 2022.
\newblock \href {https://doi.org/10.18653/v1/2022.acl-long.292} {{P}rom{DA}: Prompt-based data augmentation for low-resource {NLU} tasks}.
\newblock In \emph{Proceedings of the 60th Annual Meeting of the Association for Computational Linguistics (Volume 1: Long Papers)}, pages 4242--4255, Dublin, Ireland. Association for Computational Linguistics.

\bibitem[{Williams et~al.(2018)Williams, Nangia, and Bowman}]{williams-etal-2018-broad}
Adina Williams, Nikita Nangia, and Samuel Bowman. 2018.
\newblock \href {https://doi.org/10.18653/v1/N18-1101} {A broad-coverage challenge corpus for sentence understanding through inference}.
\newblock In \emph{Proceedings of the 2018 Conference of the North {A}merican Chapter of the Association for Computational Linguistics: Human Language Technologies, Volume 1 (Long Papers)}, pages 1112--1122, New Orleans, Louisiana. Association for Computational Linguistics.

\bibitem[{Wr{\'o}blewska and Krasnowska-Kiera{\'s}(2017)}]{wroblewska-krasnowska-kieras-2017-polish}
Alina Wr{\'o}blewska and Katarzyna Krasnowska-Kiera{\'s}. 2017.
\newblock \href {https://doi.org/10.18653/v1/P17-1073} {{P}olish evaluation dataset for compositional distributional semantics models}.
\newblock In \emph{Proceedings of the 55th Annual Meeting of the Association for Computational Linguistics (Volume 1: Long Papers)}, pages 784--792, Vancouver, Canada. Association for Computational Linguistics.

\bibitem[{Yan et~al.(2021)Yan, Li, Wang, Zhang, Wu, and Xu}]{yan-etal-2021-consert}
Yuanmeng Yan, Rumei Li, Sirui Wang, Fuzheng Zhang, Wei Wu, and Weiran Xu. 2021.
\newblock \href {https://doi.org/10.18653/v1/2021.acl-long.393} {{C}on{SERT}: A contrastive framework for self-supervised sentence representation transfer}.
\newblock In \emph{Proceedings of the 59th Annual Meeting of the Association for Computational Linguistics and the 11th International Joint Conference on Natural Language Processing (Volume 1: Long Papers)}, pages 5065--5075, Online. Association for Computational Linguistics.

\bibitem[{Yang et~al.(2021)Yang, Yang, Cer, Law, and Darve}]{yang-etal-2021-universal}
Ziyi Yang, Yinfei Yang, Daniel Cer, Jax Law, and Eric Darve. 2021.
\newblock \href {https://doi.org/10.18653/v1/2021.emnlp-main.502} {Universal sentence representation learning with conditional masked language model}.
\newblock In \emph{Proceedings of the 2021 Conference on Empirical Methods in Natural Language Processing}, pages 6216--6228, Online and Punta Cana, Dominican Republic. Association for Computational Linguistics.

\bibitem[{Zhou et~al.(2022{\natexlab{a}})Zhou, Zheng, Tang, Jian, and Yang}]{zhou-etal-2022-flipda}
Jing Zhou, Yanan Zheng, Jie Tang, Li~Jian, and Zhilin Yang. 2022{\natexlab{a}}.
\newblock \href {https://doi.org/10.18653/v1/2022.acl-long.592} {{F}lip{DA}: Effective and robust data augmentation for few-shot learning}.
\newblock In \emph{Proceedings of the 60th Annual Meeting of the Association for Computational Linguistics (Volume 1: Long Papers)}, pages 8646--8665, Dublin, Ireland. Association for Computational Linguistics.

\bibitem[{Zhou et~al.(2022{\natexlab{b}})Zhou, Lei, Liu, Du, Huang, Zhao, Dai, Chen, Le, and Laudon}]{NEURIPS2022_2f00ecd7}
Yanqi Zhou, Tao Lei, Hanxiao Liu, Nan Du, Yanping Huang, Vincent Zhao, Andrew~M Dai, zhifeng Chen, Quoc~V Le, and James Laudon. 2022{\natexlab{b}}.
\newblock \href {https://proceedings.neurips.cc/paper_files/paper/2022/file/2f00ecd787b432c1d36f3de9800728eb-Paper-Conference.pdf} {Mixture-of-experts with expert choice routing}.
\newblock In \emph{Advances in Neural Information Processing Systems}, volume~35, pages 7103--7114. Curran Associates, Inc.

\end{thebibliography}
